\documentclass[11pt]{article}
\usepackage[preprint]{acl}

\usepackage{times}
\usepackage{latexsym}
\usepackage{amsmath}
\usepackage{enumitem}
\usepackage{CJKutf8}
\usepackage{multirow}
\usepackage{amssymb}
\usepackage{booktabs}
\usepackage{colortbl}
\usepackage{caption}
\usepackage[normalem]{ulem}
\usepackage{booktabs} 
\usepackage{multirow} 
\usepackage{arydshln} %
\usepackage{float}
\usepackage{array}
\useunder{\uline}{\ul}{}
\usepackage{subcaption} 
\usepackage[T1]{fontenc}
\usepackage{adjustbox}
\usepackage{makecell}
\makeatletter
\newif\if@restonecol
\makeatother

\usepackage[linesnumbered,ruled]{algorithm2e}
\usepackage{algpseudocode}

\usepackage[utf8]{inputenc}

\usepackage{microtype}

\usepackage{inconsolata}

\usepackage{graphicx}

\usepackage{arydshln}  
\usepackage{xcolor}    

\title{A Multi-Agent Framework with Automated Decision Rule Optimization for Cross-Domain Misinformation Detection}

\author{Hui Li$^{1*}$, Ante Wang$^{1*}$, Kunquan Li$^{1}$,   Zhihao Wang$^{1}$, Liang Zhang$^{1}$,
\textbf{Delai Qiu}$^{2}$, \\\textbf{Qingsong Liu}$^{2}$,   \textbf{Jinsong Su}$^{1,3\dag}$\\
       $^1$ School of Informatics, Xiamen University,  China\\
       $^2$ Unisound Al Technology, China \\
       $^3$ Shanghai Artificial Intelligence Laboratory,  China\\
       \texttt{huilinlp@xmu.edu.cn\; jssu@xmu.edu.cn }
       }

\begin{document}
\maketitle

\begin{abstract}

Misinformation spans various domains, but detection methods trained on specific domains often perform poorly when applied to others. With the rapid development of Large Language Models (LLMs), researchers have begun to utilize LLMs for cross-domain misinformation detection. However, existing LLM-based methods often fail to adequately analyze news in the target domain, limiting their detection capabilities. More importantly, these methods typically rely on manually designed decision rules, which are limited by domain knowledge and expert experience, thus limiting the generalizability of decision rules to different domains. To address these issues, we propose a \textbf{M}ulti-\textbf{A}gent Framework for cross-domain misinformation detection with Automated Decision \textbf{R}ule \textbf{O}ptimization (MARO). Under this framework, we first employs multiple expert agents to analyze target-domain news. Subsequently, we introduce a \textit{question-reflection mechanism} that guides expert agents to facilitate higher-quality analysis. 
Furthermore, we propose a decision rule optimization approach based on carefully designed cross-domain validation tasks to iteratively enhance decision rule effectiveness across domains. Experimental results and analysis on commonly used datasets demonstrate that MARO achieves significant improvements over existing methods.
\end{abstract}

\section{Introduction}

\begin{figure*}[t] 
    \centering
    \includegraphics[width=
\textwidth]{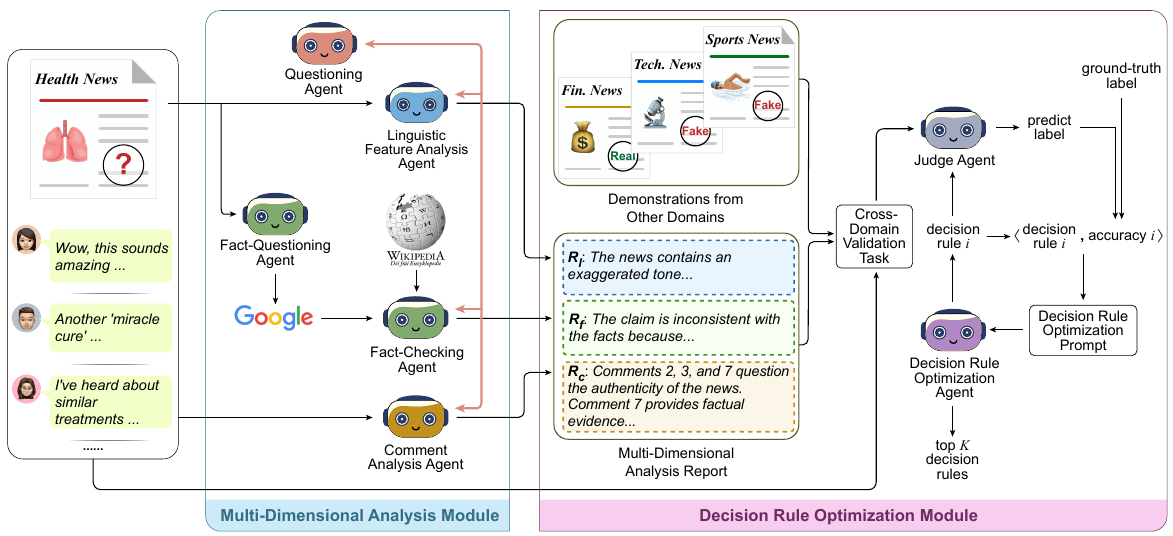} 
    \caption{{MARO first performs a multi-dimensional analysis on the news to be verified. Afterwards, the news, the multi-dimensional analysis report, and demonstration news from other domains are provided to the Judge Agent for verification. Meanwhile, the Decision Rule Optimization Agent supplies the Judge Agent with a decision rule to guide its verification. As the Judge Agent makes decision, the resulting <decision rule, accuracy> pairs form an optimization trajectory, which in turn enables the Decision Rule Optimization Agent to further refine decision rules.}}
    \label{fig1} 
    \vspace{-10px}
\end{figure*}

\begin{CJK}{UTF8}{gbsn}
Nowadays, {social media} is flooded with misinformation spanning multiple domains such as politics, economics, and technology, significantly impacting people's lives and societal stability \cite{della2023misinformation}. However, due to the differences in background knowledge and linguistic features across domains, misinformation detection models trained on specific domains often perform poorly when applied to others \cite{ran2023unsupervised, liu2024raemollm}. Thus, cross-domain misinformation detection offers substantial practical value, leading to increased research attention on this task. \cite{choudhry-etal-2022-emotion, lin-etal-2022-detect, ran2023metric, ran2023unsupervised, liu2024raemollm, karisani2024fact}.

\vspace{-0.055in}

Generally, cross-domain misinformation detection methods are trained on the mixture of multiple source-domain datasets, and then evaluated on a unseen target-domain one \cite{hernandez2017cross, lin-etal-2022-detect, ran2023metric, ran2023unsupervised}. Early studies primarily use machine learning methods with various classifiers \cite{perez2014cross, hernandez2017cross}. Subsequently, researchers resort to deep learning-based methods \cite{choudhry-etal-2022-emotion, lin-etal-2022-detect, ran2023metric, ran2023unsupervised}, which, however, suffer from limited training data. In recent years, with the emergence of Large Language Models (LLMs), researchers have shifted their attention to exploring the powerful capabilities of LLMs \cite{10480162, liu2024raemollm}. For example, \citet{10480162} explore incorporating graph knowledge into LLMs for cross-domain misinformation detection. Very recently, \citet{liu2024raemollm} propose a Retrieval-Augmented Generation approach that achieves state-of-the-art performance. They extract labeled source-domain examples based on emotional relevance and manually design a decision rule. These examples and the decision rule are incorporated into the prompt to directly judge target-domain veracity.


In spite of their success, these methods still have two major drawbacks. {First, they tend to treat misinformation detection as a monolithic task, overlooking that news understanding is inherently multi-dimensional—covering linguistic features, external factual consistency, user comments, and so on. Although \citet{wan2024dell} makes an initial attempt to incorporate multiple proxy tasks, their analysis remains inadequate\footnote{We validate this issue in Section \ref{Further Analysis} through experiments.}.} More importantly, these methods rely on manually designed decision rules, which are typically developed based on domain-specific knowledge and experts' experience. However, news from different domains often exhibit different background knowledge and linguistic features. As a result, these decision rules usually struggle to effectively detect misinformation across different domains, leading to poor adaptability.


In this paper, we propose a \textbf{M}ulti-\textbf{A}gent Framework for cross-domain misinformation detection with Automated Decision \textbf{R}ule \textbf{O}ptimization, MARO. As illustrated in Figure \ref{fig1}, MARO consists of two main modules: 1) {Multi-Dimensional Analysis Module, {which decomposes the complex analysis task into several subtasks, each handled by an expert agent focusing on a specific aspect}—such as linguistic features, external fact consistency, and user comments—collectively producing a set of analysis reports.} In particular, to improve the quality of these analyses, we introduce a question-reflection mechanism, which employs a Questioning Agent to generate corresponding reflection questions based on the initial analysis reports, thereby helping the above expert agents produce more refined analysis responses. 2) The Decision Rule Optimization Module, which is specifically designed to automatically optimize and generate more effective decision rules. For this purpose, we gather news from different domains within the source-domain dataset and construct a series of validation tasks designed to simulate cross-domain misinformation detection scenarios. This module iteratively optimizes the decision rules according to their performance on the validation tasks.

We evaluate the performance of MARO using two commonly-used cross-domain misinformation detection datasets. Experimental results show that MARO outperforms existing state-of-the-art baselines across multiple LLMs. Further experiments demonstrate that both Multi-Dimensional Analysis Module and Decision Rule Optimization Module effectively improve the performance of MARO.

\end{CJK}

\begin{CJK}{UTF8}{gbsn}

\vspace{-2px}
\section{Our Method}

\subsection{Task Formulation}

Given multiple source domain news datasets $D_s=\{D_{s}^i\}_{i=1}^{|D_s|}$ and a target domain news datasets $D_t$, each domain contains multiple news items represented as ${(x_j, c_j, y_j)}_{j=1}^{|D_*|}$, where $x_j$ denotes the news content, $c_j=\{c_j^k\}_{k=1}^{|c_j|}$ represents the set of comments related to $x_j$, and $y_j \in \{0,1\}$ is the corresponding ground-truth label. The goal of the cross-domain misinformation detection is to use source domain data to learn model parameters or decision rules with sufficient generalizability, and then effectively apply them to the target domain.

\subsection{MARO}

As shown in Figure \ref{fig1}, MARO consists of two main modules: the Multi-Dimensional Analysis Module and the Decision Rule Optimization Module, both of which employ {LLM-based} agents to perform various tasks. We provide  comprehensive details of these modules in the following subsections.

\subsubsection{Multi-Dimensional Analysis Module}

This module employs analysis agents to examine a news item from multiple perspectives, generating a multi-dimensional report to support decision making. To this end, we design four kinds of agents: \textit{Linguistic Feature Analysis Agent}, \textit{Comment Analysis Agent}, \textit{Fact-Checking Agent Group}, and \textit{Questioning Agent}. Each agent (or agent group) focuses on a specific aspect of the news item, collectively providing a comprehensive analysis report.

\vspace{5pt}

\noindent\textbf{Linguistic Feature Analysis Agent.}\quad This agent analyzes linguistic features of the news content, such as emotional tone and writing style, generating a \textit{linguistic feature analysis report} $R_l$. Specifically, we design a system prompt $P_l$ to guide the LLM in analyzing linguistic features of the news, producing the report $R_l$ as $R_l = \text{LLM}(P_l, x)$. The blue dashed box in Figure \ref{fig1} presents a simplified linguistic feature analysis report, identifying an exaggerated tone in the news content.

\vspace{5pt}

\noindent\textbf{Comment Analysis Agent.}\quad This agent analyzes comments to identify commenters' stances, emotional attitudes, and evidence information. It generates a \textit{comment analysis report} $R_c$ that summarizes commenters' reactions and factual evidence while counting their opinion distribution: $R_c = \text{LLM}(P_c, x, c)$, where \( P_c \) is the system prompt for Comment Analysis Agent. The orange dashed box in Figure \ref{fig1} offers a simplified view of the generated comment analysis report, which quantifies the distribution of commenters' opinions and presents fact evidence. 

\vspace{5pt}

\noindent\textbf{Fact-Checking-Agent Group.}\quad  This agent group uses external facts to verify the authenticity of news. It primarily consists of two agents: a \textit{Fact-Questioning Agent} and a \textit{Fact-Checking Agent}.

The Fact-Questioning Agent generates yes/no questions based on claims in the news content. The fact question set \( Q_f \) is generated as $Q_f = \text{LLM}(P_{Q_f}, x),$ where \( P_{Q_f} \) is the system prompt for Fact-Questioning Agent. Then, $Q_f$ serve as queries to retrieve relevant clues from Google.

The Fact-Checking Agent combines clues retrieved from Google and facts gathered via the Wikipedia tool to collect an evidence set $e$. Subsequently, it evaluates the consistency between claims in news content and $e$.
Based on this evaluation, it generates a fact-checking analysis report $R_f$ to identify misleading claims: $R_f = \text{LLM}(P_f, x, e)$, where \( P_f \) is the system prompt for Fact-Checking Agent. The green dashed box in Figure \ref{fig1} presents an example of the generated fact-checking analysis report, which highlights the inconsistency between claims in news content and the evidence. 

\vspace{5pt}

\noindent\textbf{Questioning Agent.}\quad    To ensure sufficient analysis, we introduce a question-reflection mechanism.  It uses a Questioning Agent to review the above-mentioned analysis reports, so as to identify any previously overlooked aspects. Then it generates specific questions to guide these analysis agents in conducting more in-depth and comprehensive analysis. Formally, the generation processes of these question sets are described as
\begin{align*}
Q_r^l &= \text{LLM}(P_q, x, R_l),  \\
Q_r^c &= \text{LLM}(P_q, x, c, R_c),\\
Q_r^f &= \text{LLM}(P_q, x, e, R_f),
\end{align*}
where \(Q_r^l, Q_r^c, Q_r^f\) represents the question sets for the linguistic feature analysis, comment analysis, and fact-checking analysis report, respectively. \( P_q \) is the system prompt for Questioning Agent.

The above question sets are respectively fed into the Linguistic Feature Analysis Agent, Comment Analysis Agent, and Fact-Checking Agent, enabling them to perform more comprehensive and in-depth analyses. Then, each agent produces its individual response. Finally, we integrate the three analysis reports and these responses into a unified multi-dimensional analysis report, which serves as a reliable basis for evaluating news authenticity. The system prompts for the Multi-Dimensional Analysis Module are provided in Appendix \ref{Appendix A.1}.

\subsubsection{Decision Rule Optimization Module}
\label{sec:decision_rule_optimization}

In this module, we design cross-domain verification tasks and use the module to perform them. Subsequently, we optimize decision rules based on feedback from these executions to improve their generalization across domains.

\begin{figure}[t]
    \centering
    \includegraphics[width=1\linewidth]{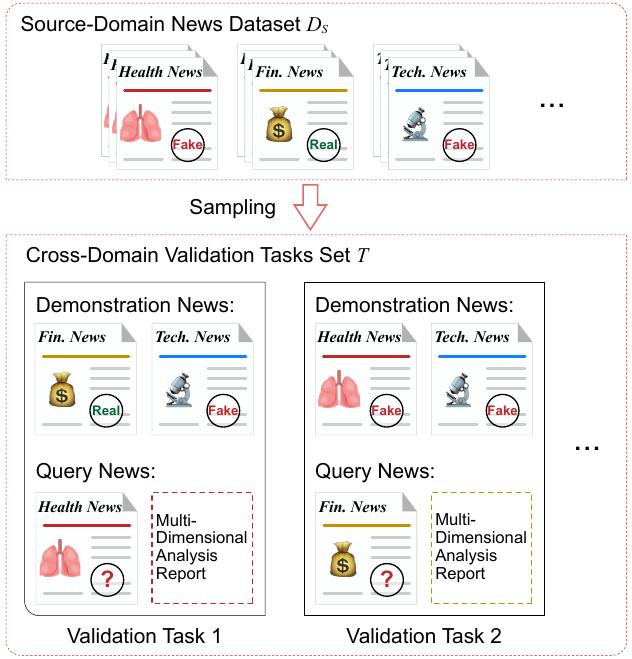}  
    \caption{An illustration of constructing cross-domain validation tasks.}
    \label{fig2} 
    \vspace{-10px}
\end{figure}

\vspace{5pt}

\noindent\textbf{Cross-Domain Validation Tasks Construction.}\quad We construct cross-domain validation tasks using news from different source domains. As illustrated in Figure \ref{fig2}, we first {randomly} sample a piece of source-domain news as the query news, and {randomly} select other source-domain annotated news as the demonstration news. The query news, along with its multi-dimensional analysis report and demonstration news, are then input into a Judge Agent in the form of in-context learning. Finally, the Judge Agent evaluates the query news and its analysis report, using the demonstration news and the decision rule to judge its truthfulness. To ensure the diversity of validation tasks, we sequentially sample query news from each source domain, thereby creating a set of cross-domain validation tasks \( T = \{t_1, t_2, \dots, t_{N_{ct}}\} \), where \( N_{ct} \) denotes the total number of cross-domain validation tasks.

\vspace{5pt}

\noindent\textbf{Decision Rule Optimization.}\quad 
To optimize the decision rules, we introduce a \textit{Decision Rule Optimization Agent}, which refines decision rules based on the feedback obtained from Judge Agent's execution on the cross-domain validation task set. As illustrated in Algorithm \ref{algorithm}, we first manually define a decision rule \(r_0\). Using $r_0$, the Judge Agent executes cross-domain validation task set $T$ to produce judgements. These judgements are compared with the ground-truth labels to obtain an accuracy score $s_0$. Subsequently, we add \(\langle r_0, s_0 \rangle\) to \(L_{RS}\), a set designed to store \(\langle \text{decision rule}, \text{accuracy} \rangle\) pairs (\textbf{Lines 1-2}). Furthermore, \(\langle r_0, s_0 \rangle\) is added to the optimization trajectory used in the Decision Rule Optimization Agent's prompt \(P_o\) (\textbf{Line 3}), which is provided in Appendix \ref{Appendix A.2}.

\begin{algorithm}[!t]
\caption{Decision Rule Optimization}
\label{algorithm} 
\KwIn{
\begin{itemize}[label={}, noitemsep, topsep=0pt]
    \item $T$: cross-domain validation task set
    \item $r_0$: manually defined initial decision rule 
    \item $N_{iter}$: the maximum number of iterations 
    \item $N_{att}$: the maximum number of attempts
    \item $K$: the number of returned decision rules
\end{itemize}
}

The Judge Agent utilizes $r_0$ to execute $T$, obtaining the accuracy $s_0$

$L_{RS} \leftarrow L_{RS} \cup \langle r_0, s_0 \rangle$

Add $\langle r_0, s_0 \rangle$ to the optimization trajectory

$r_{best}, s_{max}  \leftarrow r_0, s_0$

$n_{iter}, n_{att} \leftarrow 0$

\While{$n_{iter}< N_{iter}$ \textbf{and} $n_{att} < N_{att}$}{
    $n_{iter}$ = $n_{iter}$ + 1

    The Decision Rule Optimization Agent generates a new decision rule $r_i$

    The Judge Agent utilizes $r_i$ to execute $T$, obtaining the accuracy $s_i$

    \eIf{$s_i > s_{\text{max}}$}{
    $L_{RS} \leftarrow L_{RS} \cup \langle r_i, s_i \rangle$
        
        $r_{{best}}, s_{{max}} \leftarrow r_i, s_i$
         
        $n_{att} \leftarrow 0$
    }{
        $n_{att}$ = $n_{att}$ + 1
    }

    Use the top 10 $\langle \text{decision rule}, \text{accuracy} \rangle$ pairs in $L_{RS}$ to construct the optimization trajectory in $P_o$
}
\Return top $K$ decision rules
\end{algorithm}

\vspace{-2px}

\begin{table*}[hbt]
\centering
 \renewcommand\arraystretch{1} 
\small
\setlength\tabcolsep{4pt}
\begin{tabular}{cl|ccc ccccc cc }
\toprule
\multicolumn{2}{c|}{\multirow{2}{*}{\bf Method}}&\multicolumn{2}{c}{\bf Disasters}&\multicolumn{2}{c}{\bf Entertain}&\multicolumn{2}{c}{\bf Health}&\multicolumn{2}{c}{\bf Politics}&\multicolumn{2}{c}{\bf Society} \\ \cmidrule(l){3-4}\cmidrule(l){5-6}\cmidrule(l){7-8}\cmidrule(l){9-10}\cmidrule(l){11-12}
&&\bf Acc.&\bf F1&\bf Acc.&\bf F1&\bf Acc.&\bf F1&\bf Acc.&\bf F1&\bf Acc.&\bf F1 \\ \midrule
\multirow{3}{*}{NN-based}&UCD-RD \cite{ran2023unsupervised}&70.26&69.94&56.05&56.57&70.9&71.35&62.19&61.78&61.09&60.95\\
&CADA \cite{li2023improving}&73.26&72.75&58.24&58.05&70.3&70.05&64.33&65.07&59.82&58.62\\
&ADAF \cite{karisani2024fact} &73.54&72.39&57.19&56.95&70.5&69.91&62.82&61.94&61.19&61.88\\  \midrule
& \vspace*{-0.2cm}\\
\multirow{8}{*}{LLM-based}&GPT-3.5  w/ tools &72.35&72.19&60.26&59.91&68.8&68.05&63.42&62.94&61.69&60.27\\
&HiSS \cite{zhang2023towards}&72.79&72.06&57.56&56.87&72.7&72.37&67.96&66.34&62.64&61.07\\
&SAFE \cite{wei2024long}&71.84&70.97&60.75&60.37&71.9&70.07&65.04&64.32&61.67&60.28\\
&TELLER \cite{liu-etal-2024-teller} &75.28&74.67&60.28&60.57&75.2&74.86&65.18&64.97&63.57&63.87\\
&DELL \cite{wan2024dell} &75.26&74.05&{\ul 65.67}&{\ul 64.95}&76.1&75.81&67.59&66.95&63.82&63.39\\
&DeepSeek-R1 \cite{deepseekai2025deepseekr1incentivizingreasoningcapability} &76.41&{\ul 85.73}&57.16&54.86&69.5&76.3&{\ul 72.15}&\bf 80.11&{\ul 66.89}&\bf 72.19\\
&RAEmo \cite{liu2024raemollm} &{\ul 78.29}& 78.84&61.51&60.37&{\ul 77.3}&{\ul 76.87}& 68.74&70.87&64.78&65.06\\
&\cellcolor{gray!50}MARO (ours)&\cellcolor{gray!50}\bf 82.98&\cellcolor{gray!50}\bf 88.15&\cellcolor{gray!50}\bf 67.54& \cellcolor{gray!50}\textbf{65.9}&\cellcolor{gray!50}\bf 81.9&\cellcolor{gray!50}\bf 82.37&\cellcolor{gray!50}\bf 74.97& \cellcolor{gray!50}{\ul 79.38}&\cellcolor{gray!50}\bf 69.96&\cellcolor{gray!50}{\ul 71.97}\\ \midrule\midrule
\multicolumn{2}{c|}{\multirow{2}{*}{\bf Method}}&\multicolumn{2}{c}{\bf Education}&\multicolumn{2}{c}{\bf Finance}&\multicolumn{2}{c}{\bf Military}&\multicolumn{2}{c}{\bf Science}&\multicolumn{2}{c}{\bf Avg.} \\ \cmidrule(l){3-4}\cmidrule(l){5-6}\cmidrule(l){7-8}\cmidrule(l){9-10}\cmidrule(l){11-12}
&&\bf Acc.&\bf F1&\bf Acc.&\bf F1&\bf Acc.&\bf F1&\bf Acc.&\bf F1&\bf Acc.&\bf F1 \\ \midrule
\multirow{3}{*}{NN-based}&UCD-RD \cite{ran2023unsupervised}&60.84&60.74&60.11&59.89&69.05&69.47&57.58&57.32&63.12&63.11\\
&CADA \cite{li2023improving}&64.31&63.82&61.15&60.83&69.37&70.14&59.31&59.14&64.45&64.27\\
&ADAF \cite{karisani2024fact} &65.54&64.32&62.05&61.16&66.28&65.16&59.16&58.49&64.25&63.58\\ 
\midrule 
& \vspace*{-0.2cm}\\
\multirow{8}{*}{LLM-based}&GPT-3.5  w/ tools &65.96&65.79&62.15&61.61&67.41&66.27&60.16&59.65&64.69&64.08\\
&HiSS \cite{zhang2023towards}&64.84&64.15&63.95&62.89&68.63&67.84&55.91&55.37&65.22&64.33\\
&SAFE \cite{wei2024long}&64.95&64.12&60.56&60.13&68.21&68.14&57.73&56.65&64.74&63.89\\
&TELLER \cite{liu-etal-2024-teller} &67.79&67.08&{\ul 65.06}&{\ul 65.27}&71.05&70.39&60.05&59.89&67.05&66.84\\
&DELL \cite{wan2024dell} &69.05&68.31&62.65&63.49&67.26&66.86&59.76&58.14&67.46&66.88\\
&DeepSeek-R1 \cite{deepseekai2025deepseekr1incentivizingreasoningcapability} &62.19&{\ul 72.07}&57.26&55.26&{\ul 73.82}&{\ul 79.04}&57.38&{\ul 62.17}& 65.87&{\ul 70.86}\\
&RAEmo \cite{liu2024raemollm} &{\ul 69.26}&70.73&64.25&63.53&72.86& 71.49&{\ul 60.63}&60.17&{\ul 68.62}&68.66\\
&\cellcolor{gray!50}MARO (ours)&\cellcolor{gray!50}\bf 74.79&\cellcolor{gray!50}\bf 75.1&\cellcolor{gray!50}\bf 71.48& \cellcolor{gray!50}\textbf{67.97}&\cellcolor{gray!50}\bf 81.12&\cellcolor{gray!50}\bf 84.97&\cellcolor{gray!50}\bf 66.66&\cellcolor{gray!50}\textbf{63.82}&\cellcolor{gray!50}\bf 74.6&\cellcolor{gray!50}\bf 75.51
\\ \bottomrule
\end{tabular}
\caption{Performance comparison between MARO and baselines on Weibo21 using GPT-3.5-turbo-0125 as the underlying model. NN-based denotes conventional neural network-based methods. GPT-3.5  w/ tools means  we enable GPT-3.5-turbo-0125 to make independent judgments using the search engine and the Wikipedia tool. The best result in each column is marked in \textbf{bold} and the second best is {\ul underlined}. All results are reported as percentages.}
\label{table weibo21 main}
\vspace{-2px}
\end{table*}

We design an iterative optimization process to progressively enhance the generalizability of generated decision rules (\textbf{Lines 6-16}). During each iteration, the Decision Rule Optimization Agent first generates a new decision rule \(r_i\), which is then applied by the Judge Agent to the cross-domain validation task set \(T\) (\textbf{Lines 8-9}). If $s_i$ exceeds $s_{max}$, the pair \(\langle r_i, s_i \rangle\) is added to \(L_{RS}\), and we update the best decision rule \(r_{best}\), the maximum accuracy \(s_{max}\) with \(r_i\) and  \(s_i\) (\textbf{Lines 11-12}). Next, we select the top 10 \(\langle \text{decision rule}, \text{accuracy} \rangle\) pairs from \(L_{RS}\) to update the optimization trajectory in \(P_o\) (\textbf{Line 17}). This enables the Decision Rule Optimization Agent to iteratively refine decision rules, ultimately achieving higher accuracy. Through this process, we expand \(L_{RS}\) until reaching the maximum iteration limit \(N_{iter}\) or failing to surpass \(s_{max}\) for \(N_{att}\) consecutive iterations (\textbf{Line 6}). Finally, the Decision Rule Optimization Module outputs the top \(K\) decision rules from \(L_{RS}\) (\textbf{Line 19}).

\subsubsection{{Inference}}

{During inference, the news and its multi-dimensional analysis report are provided to the Judge Agent, which evaluates the input using each of the top \(K\) optimized decision rules. The final judgement is determined by majority voting.}

\end{CJK}


%

\begin{table*}[hbt]
\centering
\small
\setlength\tabcolsep{6.75pt}
\begin{tabular}{cl|ccc ccccc}
\toprule
\multicolumn{2}{c|}{\multirow{2}{*}{\bf Method}}&\multicolumn{2}{c}{\bf Biz}&\multicolumn{2}{c}{\bf Edu}&\multicolumn{2}{c}{\bf Cele}&\multicolumn{2}{c}{\bf Entmt}\\ \cmidrule(l){3-4}\cmidrule(l){5-6}\cmidrule(l){7-8}\cmidrule(l){9-10}
&&\bf Acc.&\bf F1&\bf Acc.&\bf F1&\bf Acc.&\bf F1&\bf Acc.&\bf F1\\ \midrule
\multirow{3}{*}{NN-based}&UCD-RD \cite{ran2023unsupervised}&73.52&73.29&64.21&63.85&62.2&61.93&61.57&60.21\\
&CADA \cite{li2023improving}&74.33&74.62&66.98&66.55&62&60.63&60.95&59.94\\
&ADAF \cite{karisani2024fact} &78.62&77.85&70.82&70.71&63.8&62.83&62.82&62.95\\   \midrule
& \vspace*{-0.2cm}\\
\multirow{8}{*}{LLM-based}& GPT-3.5  w/ tools&80.17&80.51&72.19&71.07&64.6&62.06&62.12&58.01\\
&HiSS \cite{zhang2023towards}&77.13&77.48&72.57&71.06&66.4&{\ul 66.79}&62.58&61.84\\
&SAFE \cite{wei2024long}&79.26&78.64&72.51&72.27&63.8&62.11&63.56&63.13\\
&TELLER \cite{liu-etal-2024-teller} &82.21&81.38&73.27&73.85&{\ul 67.6}&65.28&{\ul 63.91}&{\ul 63.64}\\
&DELL \cite{wan2024dell} &{\ul 83.57}&{\ul 82.94}&{\ul 74.13}&73.72&65.2&64.35&62.54&61.49\\
&DeepSeek-R1 \cite{deepseekai2025deepseekr1incentivizingreasoningcapability} &82.5&81.57&71.25&{\ul 74.15}&65&65.34&63.75&61.33\\
&RAEmo \cite{liu2024raemollm} &78.76&77.16&69.28&68.07&61&59.27&61.13&60.21\\
 &\cellcolor{gray!50}MARO (ours)&\cellcolor{gray!50}\bf 85.46&\cellcolor{gray!50}\bf 84.83&\cellcolor{gray!50}\bf 77.62&\cellcolor{gray!50}\bf 77.24&\cellcolor{gray!50}\bf 68.8&\cellcolor{gray!50}\bf 67.95&\cellcolor{gray!50}\bf 66.81&\cellcolor{gray!50}\bf 65.97\\ \midrule\midrule
\multicolumn{2}{c|}{\multirow{2}{*}{\bf Method}}&\multicolumn{2}{c}{\bf Polit}&\multicolumn{2}{c}{\bf Sport}&\multicolumn{2}{c}{\bf Tech}&\multicolumn{2}{c}{\bf Avg.}\\ \cmidrule(l){3-4}\cmidrule(l){5-6}\cmidrule(l){7-8}\cmidrule(l){9-10}
&&\bf Acc.&\bf F1&\bf Acc.&\bf F1&\bf Acc.&\bf F1&\bf Acc.&\bf F1\\ \midrule
\multirow{3}{*}{NN-based}&UCD-RD \cite{ran2023unsupervised}&66.25&66.35&63.56&62.79&73.26&73.39&66.37&65.97\\
&CADA \cite{li2023improving}&68.41&68.92&63.82&62.91&72.19&73.05&66.95&66.66\\
&ADAF \cite{karisani2024fact} &71.72&71.45&71.72&71.24&72.73&72.42&70.32&69.92\\   \midrule
& \vspace*{-0.2cm}\\
\multirow{8}{*}{LLM-based}&GPT-3.5  w/ tools&71.07&73.71&72.72&70.51&74.45&75.28&71.05&70.16\\
&HiSS \cite{zhang2023towards}&71.66&70.95&74.32&73.43&72.54&71.21&71.03&70.39\\
&SAFE \cite{wei2024long}&74.51&74.76&70.75&69.63&76.51&75.86&71.56&70.91\\
&TELLER \cite{liu-etal-2024-teller} &73.57&72.29&75.24&75.51&76.11&75.65&73.13&72.51\\
&DELL \cite{wan2024dell} &{\ul 75.26}&{\ul 75.18}&\bf 79.82&{\ul 78.56}&{\ul 77.63}&{\ul 76.41}&{\ul 74.02}&{\ul 73.24}\\
&DeepSeek-R1 \cite{deepseekai2025deepseekr1incentivizingreasoningcapability} &65&65.85&71.25&68.49&71.25&70.12&70.00&69.55\\
&RAEmo \cite{liu2024raemollm} &73.55&72.58&71.15&70.09&70.89&69.05&69.39&68.06\\ 
&\cellcolor{gray!50}MARO (ours)&\cellcolor{gray!50}{\bf78.93}&\cellcolor{gray!50}\bf 79.73&\cellcolor{gray!50}{\ul 79.65}&\cellcolor{gray!50}\bf 79.34&\cellcolor{gray!50}\bf 82.86&\cellcolor{gray!50}\bf 82.47&\cellcolor{gray!50}\bf 77.16&\cellcolor{gray!50}\bf 76.79
\\ \bottomrule
\end{tabular}
\caption{Performance comparison between MARO and the baselines on AMTCele.}
\vspace{-10px}
\label{table amt main}
\end{table*}

\section{Experiments}

\subsection{Setup}

\noindent\textbf{Datasets.}\quad    We conduct experiments on the Weibo21 \cite{nan2021mdfend} and AMTCele \cite{liu2024raemollm} datasets. Weibo21 is a {Chinese} multi-domain rumor detection dataset covering 9 domains, where each news item includes news content and several comments. AMTCele, constructed by \citet{liu2024raemollm}, is an {English} fake news detection dataset covering 7 domains. In this dataset, each news item contains only news content. Further details are provided in Appendix \ref{datasets details}.


\vspace{5pt}

\noindent\textbf{Baselines.}\quad   We compare MARO with two kinds of baselines: 1) \textbf{conventional neural networks based methods}:  UCD-RD \cite{ran2023unsupervised}, CADA \cite{li2023improving} and ADAF \cite{karisani2024fact}; 2) \textbf{LLM-based methods}: HiSS \cite{zhang2023towards}, SAFE \cite{wei2024long}, TELLER \cite{liu-etal-2024-teller}, DELL \cite{wan2024dell},  DeepSeek-R1 \cite{deepseekai2025deepseekr1incentivizingreasoningcapability} and RAEmo \cite{liu2024raemollm}. Appendix \ref{baselines} provides a detailed description of these baselines.

\vspace{5pt}

\noindent\textbf{Settings and Evaluation.}\quad 
To ensure fair comparisons, we use the same underlying models to construct MARO and LLM-based baselines. Particularly, we set the temperature of the Decision Rule Optimization Agent to 1 to encourage greater diversity in outputs, and set the temperature of the Judge Agent to 0 for consistent outputs. In our experiments, we conduct 8-fold cross-validation on Weibo21 and 6-fold cross-validation on AMTCele, setting the cross-domain validation task number $N_{vt}$ to 500 for Weibo21 and 400 for AMTCele, with results shown in Appendix \ref{Cross-validation Experiments}. For both datasets, we empirically set the number of samples for each source domain to 100 on Weibo21 and 80 on AMTCele, the maximum iteration number $N_{iter}$ to 500 for Weibo21, the maximum attempt number $N_{att}$ to 10,  and the returned decision rule number $K$ to 3. Finally, we use accuracy (Acc.) and F1-score (F1) as evaluation metrics.  

\subsection{Main Results}

Tables \ref{table weibo21 main} and \ref{table amt main} present experimental results on Weibo21 and AMTCele\footnote{Additional experimental results are provided in Appendix, including those of MARO and baselines on other underlying models (Appendix \ref{more underlying models}), results on more datasets (Appendix \ref{More Datasets}) and efficiency comparison (Appendix \ref{Efficiency Comparison}).}. Overall, MARO achieves the best performance across most domains on both datasets. On Weibo21, MARO outperforms the second-best method, RAEmo, by 5.98 in average accuracy and 6.85 in average F1. On AMTCele, MARO surpasses the second-best method, DELL, by 3.14 in average accuracy and 3.55 in average F1. These results demonstrate the effectiveness of MARO in cross-domain misinformation detection.

\subsection{Further Analysis}
\label{Further Analysis}
\begin{table}[t!]
\centering
\small
\setlength\tabcolsep{0.4cm}
\begin{tabular}{@{}lcccc@{}}
\toprule
\multirow{2}{*}{} & \multicolumn{2}{c}{\textbf{Weibo21}}    & \multicolumn{2}{c}{\hspace*{0.3cm}\textbf{AMTCele}}     \\ \cmidrule(l){2-5} 
                  & \textbf{Acc.}          & \textbf{F1}             & \textbf{Acc.}           & \textbf{F1}             \\ \midrule
\multicolumn{1}{@{}l}{MARO}             & \textbf{74.60} & \textbf{75.51} & \textbf{77.16} & \textbf{76.79} \\
\quad\textit{w/o} LFAA          & 72.11         & 73.39          & 72.96          & 72.41          \\
\quad\textit{w/o} CAA           & 71.65         & 72.34          & -              & -              \\
\quad\textit{w/o} FCAG          & 72.38         & 73.56          & 72.62          & 71.83          \\ \quad\textit{w/o} QA            & 72.56         & 73.48          & 74.26          & 73.95          \\
\quad\textit{w/o} CDVT          & 70.21         & 71.75          & 73.27          & 72.86          \\
\quad\textit{w/o} DROA          & 69.47         & 71.62         & 72.18          & 71.75          \\ \bottomrule
\end{tabular}
\caption{{Ablation studies.}} 
\vspace{-15px}
\label{table ablation}
\end{table}
        
     

    
    
    

\noindent\textbf{Ablation Study.} \quad To verify the contributions of different components in MARO, we report the performance of MARO when these components are removed separately. Here, the components we considering include the Linguistic Feature Analysis Agent, the Comment Analysis Agent, the Fact-Checking-Agent Group, the Questioning Agent, the Cross-Domain Validation Tasks, and the Decision Rule Optimization Agent. To facilitate the subsequent descriptions, we name the variants of MARO removing different components as \textit{w/o} LFAA, \textit{w/o} CAA, \textit{w/o} FCAG, \textit{w/o} QA, \textit{w/o} CDVT and \textit{w/o} DROA, respectively.

From Table \ref{table ablation}, we can clearly find that the removal of these components leads to a performance drop, indicating the effectiveness of these components. In particular, the performance of \textit{w/o} QA shows a noticeable decline. This demonstrates that single-pass analysis is inadequate, while also proving that the question-reflection mechanism we proposed helps in identifying misinformation. 

\begin{figure}[t]
    \centering
    \includegraphics[width=0.4\textwidth]{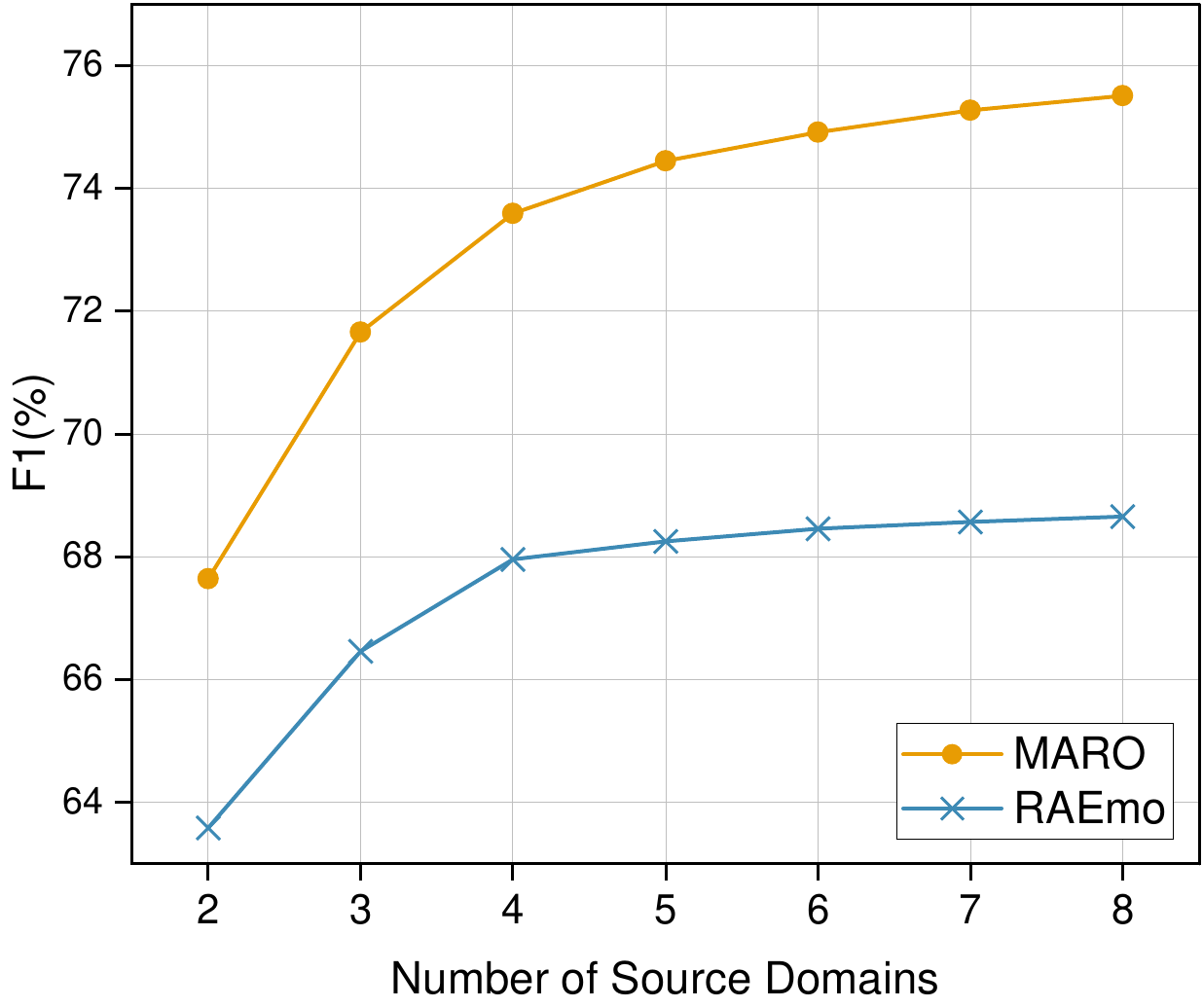}
    \caption{F1 changes with different number of source domains on Weibo21.}
    \vspace{-10px}
    \label{fig3}
\end{figure}



\vspace{5pt}

\noindent\textbf{Impact of Source Domain Number.}\quad In this experiment, we investigate how the number of source domains impacts MARO's performance. We also illustrate the performance of RAEmo, the most competitive baseline, as reported in Table \ref{table weibo21 main}.

As shown in Figure \ref{fig3}, increasing the number of source domains improves both methods' performance. This is reasonable because more source domains not only provide diverse feedback to optimize MARO’s decision rules, but also enrich RAEmo’s demonstration database. We further observe that MARO consistently outperforms RAEmo under different source domain settings, demonstrating MARO's effectiveness.

\vspace{5pt}

\noindent\textbf{Impact of Source Domain Sample Number.}\quad  Then, we investigate how the number of source domain samples affects MARO's performance. To this end, we gradually vary from 10 to 100 with an increment of 10 in each step, and report the corresponding model performance.

\begin{figure}[t]
    \centering
    \includegraphics[width=0.4\textwidth]{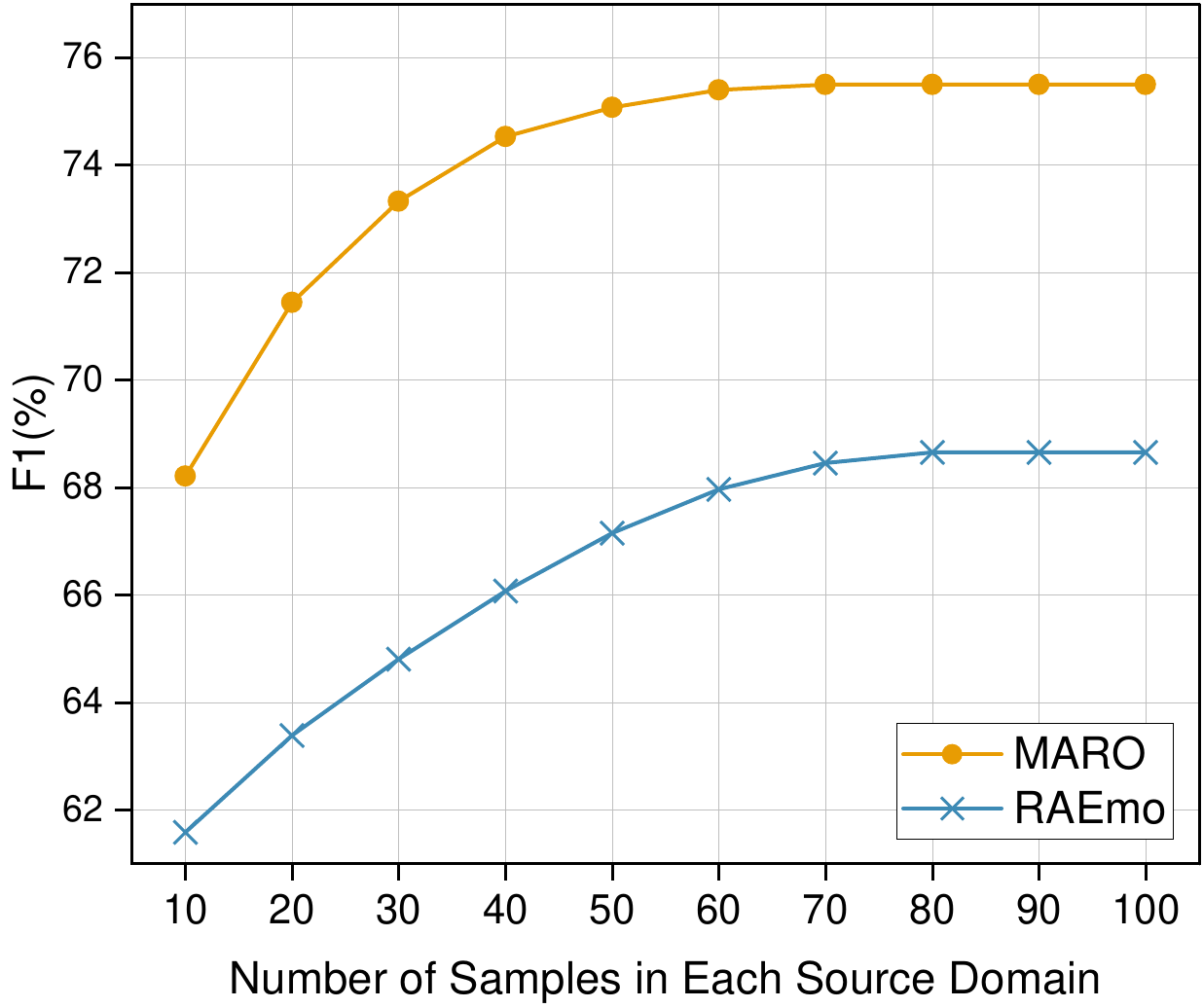}
    \caption{F1 changes with different number of samples in each source domains on Weibo21.}
    \vspace{0px}
    \label{fig4}
\end{figure}

As shown in Figure \ref{fig4}, we observe that as the number of source domain samples increases, both MARO and RAEmo show improvements in F1 scores. For this phenomena, we argue that more source-domain samples also provide more comprehensive feedback and similar demonstrations for MARO and RAEmo, respectively. Furthermore, MARO outperforms RAEmo across different numbers of source domain samples, especially in the scenarios of limited samples.




\begin{figure}[t]
    \centering
    \includegraphics[width=0.4\textwidth]{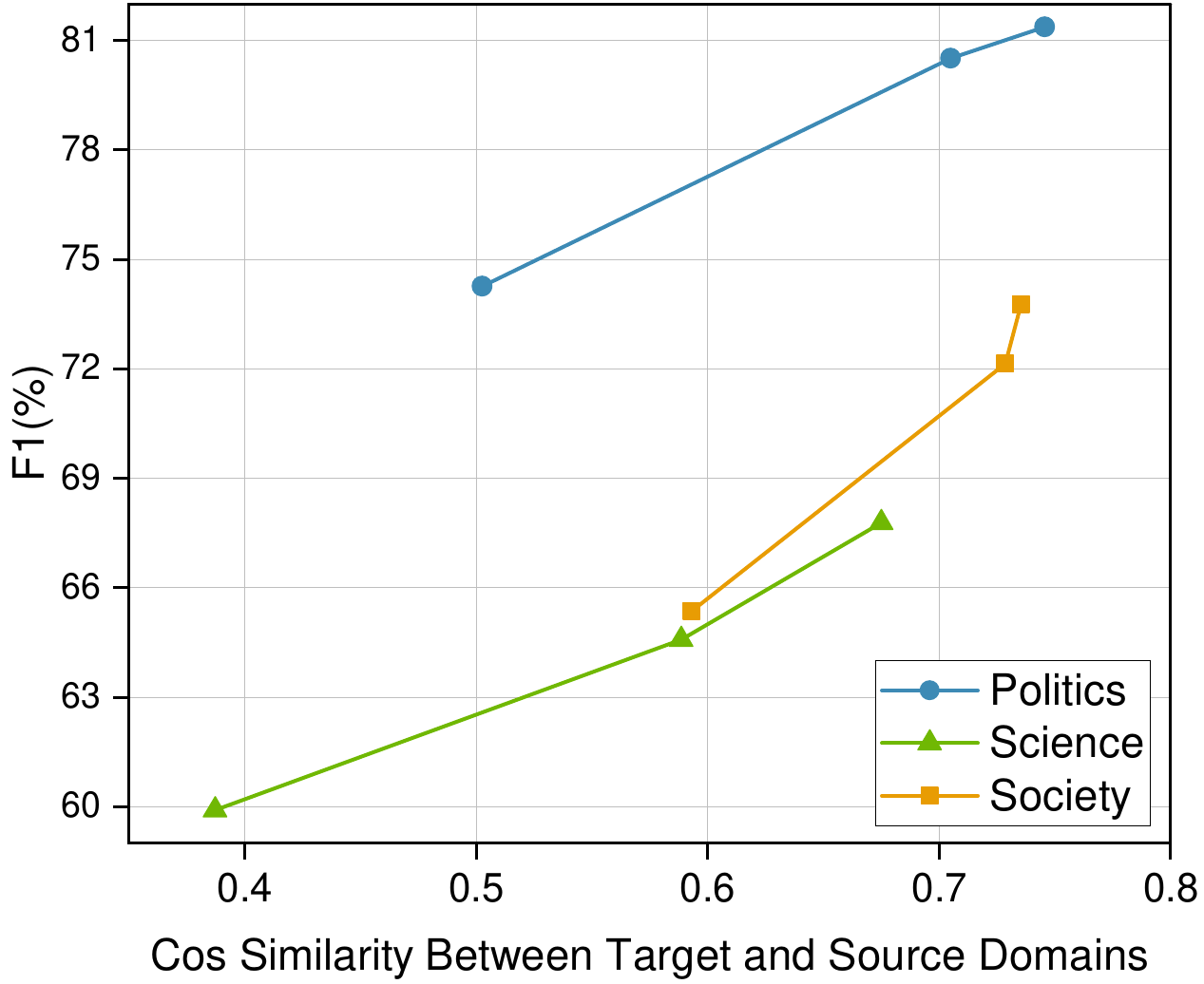}
    \caption{F1 changes with different source-target similarities on the Politics, Science and Society domains.}
    \label{F1-sim}
    \vspace{-5px}
\end{figure}

\noindent\textbf{Impact of Domain Similarity.}\quad As mentioned previously, MARO is proposed to address cross-domain misinformation detection. Thus, one critical question arises regarding the impact of the similarity between source and target domains on the performance of MARO. To investigate this, we use TF-IDF to calculate the semantic similarity between news from different domains in Weibo21, as illustrated by the similarity matrix in Appendix \ref{sim matrix}. We sample \textit{Politics}, \textit{Science}, and \textit{Society} as target domains, and pair the remaining six domains into three groups as source domains. Figure \ref{F1-sim} illustrates the relationship between source-target domain similarity and the performance of MARO. 

It can be observed from Figure \ref{F1-sim} that the performance of MARO reflects a positive correlation with domain similarity. This phenomena is reasonable since similar source domain can provide abundant shared features, which enable the Decision Rule Optimization Agent to generate decision rules that are more effective for the target domain.



\section{Case Study}

We provide an example of the decision rule optimization process in Appendix \ref{case stydy}.

\section{Related Work}

Recently, LLMs have demonstrated impressive performance across a range of tasks \cite{wang2025don, wang2025litesearch} and have been extensively used for misinformation detection \cite{huang2023harnessing, zhang2023towards, SheepDog, poblete2023using, liu2024skepticism, liu2024stepwise, liu-etal-2024-teller, wei2024long, liu2024detect, nan2024let, wan2024dell}. For example, \citet{huang2023harnessing} design prompts tailored to the features of fake news, effectively guiding ChatGPT for misinformation detection. Along this line, \citet{zhang2023towards} and \citet{wei2024long} propose to deconstruct complex claims into simpler sub-statements, which are then verified step-by-step using external search engines. Unlike the above studies, \citet{SheepDog} leverage LLMs to disguise news styles and employ style-agnostic training, thereby improving the robustness of misinformation detection systems against style variations. \citet{liu2024detect} leverage LLMs to extract key information and integrate both the model's internal knowledge and external real-time information to conduct a comprehensive multi-perspective evaluation. To address the problem of scarce comments in the early stages of misinformation spread, \citet{nan2024let} utilize LLMs to simulate users and generate diverse comments. Slightly similar to ours, \citet{wan2024dell} propose DELL, which analyzes various aspects of news to assist in identifying misinformation. Despite their effectiveness, these studies mainly concentrate on in-domain detection and have yet to adequately address the challenges of cross-domain detection.

Early approaches to cross-domain misinformation detection \cite{perez2014cross, hernandez2017cross} rely on handcrafted features and traditional models, leading to limited performance. With the advent of deep learning, researchers explore this task by aligning feature representations across domains \cite{choudhry-etal-2022-emotion} or capturing invariant features \cite{ran2023metric, ran2023unsupervised} or reducing inter-domain discrepancies \cite{lin-etal-2022-detect}. Nevertheless, the lack of sufficient cross-domain labeled data limits the effectiveness of these methods. Very recently, \citet{liu2024raemollm} propose RAEmo, which leverages an emotion-aware LLM to encode source-domain samples and create in-context learning tasks for target-domain misinformation detection. However, RAEmo still relies on manually-designed decision rules for reasoning.

We introduce a multi-dimensional analysis approach within our framework to assist in news veracity evaluation, which has not been explored in previous studies. The one exception is DELL. However, unlike DELL, we introduce a Questioning Agent to facilitate more in-depth and comprehensive analysis. More importantly, compared with studies on LLM-based misinformation detection, such as DELL and RAEmo, we incorporate a decision rule optimization module to automatically optimize decision rules, inspired by \cite{pryzant2023automatic, xu2023wizardlm, yang2024zhongjing}.


\section{Conclusion and Future Work}

In this work, we have proposed MARO, a cross-domain misinformation detection framework which addresses two key shortcomings of existing LLM-based methods: inadequate analysis and reliance on manually designed decision rules. First, MARO employs multiple expert agents to analyze news from various dimensions and generate initial analysis reports. Then, a Questioning Agent then reviews each report and poses specific questions to prompt more in-depth and comprehensive analyses. These reports and the agents’ responses are aggregated into a multi-dimensional analysis report to assist judgment. Additionally, we propose a decision rule optimization method that automatically refines decision rules based on feedback from cross-domain validation tasks. Compared to state-of-the-art methods, MARO achieves significantly higher accuracy and F1 scores on the commonly used datasets. Ablation studies confirm the effectiveness of each component.

As future work, we plan to incorporate logical reasoning and knowledge graph reasoning to conduct a deeper analysis, and to perform a more comprehensive evaluation of decision rules, thereby providing stronger evidence for their optimization. Moreover, our multi-agent coordination approach shows promising generalization potential and can be applied to other NLP tasks, such as machine translation \cite{zeng2019iterative}, text generation \cite{su2019exploring}, and style transfer \cite{zhou2020exploring}, thus demonstrating its applicability across tasks.


\section*{Limitations}

Although MARO has demonstrated effectiveness in cross-domain misinformation detection, it may have two limitations. First, MARO's workflow is complex, requiring multiple rounds of iteration to generate effective decision rules, as well as multi-dimensional analysis conducted through multiple agents. Second, the clues gathered via search engines may include misinformation fabricated by malicious actors, which may introduce distortion into the process of judging the authenticity of target-domain news.

\section*{Acknowledgement}

The project was supported by 
National Key R\&D Program of China (No. 2022ZD0160501), 
Natural Science Foundation of Fujian Province of China (No. 2024J011001),
and
the Public Technology Service Platform Project of Xiamen (No.3502Z20231043).
We also thank the reviewers for their insightful comments.
\bibliography{custom}

\begin{thebibliography}{33}
\providecommand{\natexlab}[1]{#1}

\bibitem[{Buntain and Golbeck(2017)}]{buntain2017automatically}
Cody Buntain and Jennifer Golbeck. 2017.
\newblock Automatically identifying fake news in popular twitter threads.
\newblock In \emph{SmartCloud}.

\bibitem[{Choudhry et~al.(2022)Choudhry, Khatri, Chakraborty, Vishwakarma, and Prasad}]{choudhry-etal-2022-emotion}
Arjun Choudhry, Inder Khatri, Arkajyoti Chakraborty, Dinesh Vishwakarma, and Mukesh Prasad. 2022.
\newblock Emotion-guided cross-domain fake news detection using adversarial domain adaptation.
\newblock In \emph{ICNLP}.

\bibitem[{Della~Giustina(2023)}]{della2023misinformation}
Nicholas Della~Giustina. 2023.
\newblock \emph{Misinformation and Its Effects on Individuals and Society from 2015-2023: A Mixed Methods Review Study}.
\newblock University of Washington.

\bibitem[{Guo et~al.(2025)Guo, Yang, Zhang, Song, Zhang, Xu, Zhu, Ma, Wang, Bi, Zhang, Yu, Wu, Wu, Gou, Shao, Li, Gao, Liu, Xue, Wang, Wu, Feng, Lu, Zhao, Deng, Zhang, Ruan, Dai, Chen, Ji, Li, Lin, Dai, Luo, Hao, Chen, Li, Zhang, Bao, Xu, Wang, Ding, Xin, Gao, Qu, Li, Guo, Li, Wang, Chen, Yuan, Qiu, Li, Cai, Ni, Liang, Chen, Dong, Hu, Gao, Guan, Huang, Yu, Wang, Zhang, Zhao, Wang, Zhang, Xu, Xia, Zhang, Zhang, Tang, Li, Wang, Li, Tian, Huang, Zhang, Wang, Chen, Du, Ge, Zhang, Pan, Wang, Chen, Jin, Chen, Lu, Zhou, Chen, Ye, Wang, Yu, Zhou, Pan, Li, Zhou, Wu, Ye, Yun, Pei, Sun, Wang, Zeng, Zhao, Liu, Liang, Gao, Yu, Zhang, Xiao, An, Liu, Wang, Chen, Nie, Cheng, Liu, Xie, Liu, Yang, Li, Su, Lin, Li, Jin, Shen, Chen, Sun, Wang, Song, Zhou, Wang, Shan, Li, Wang, Wei, Zhang, Xu, Li, Zhao, Sun, Wang, Yu, Zhang, Shi, Xiong, He, Piao, Wang, Tan, Ma, Liu, Guo, Ou, Wang, Gong, Zou, He, Xiong, Luo, You, Liu, Zhou, Zhu, Xu, Huang, Li, Zheng, Zhu, Ma, Tang, Zha, Yan, Ren, Ren, Sha, Fu, Xu, Xie, Zhang, Hao, Ma, Yan, Wu, Gu,
  Zhu, Liu, Li, Xie, Song, Pan, Huang, Xu, Zhang, and Zhang}]{deepseekai2025deepseekr1incentivizingreasoningcapability}
Daya Guo, Dejian Yang, Haowei Zhang, Junxiao Song, Ruoyu Zhang, Runxin Xu, Qihao Zhu, Shirong Ma, Peiyi Wang, Xiao Bi, Xiaokang Zhang, Xingkai Yu, Yu~Wu, Z.~F. Wu, Zhibin Gou, Zhihong Shao, Zhuoshu Li, Ziyi Gao, Aixin Liu, Bing Xue, Bingxuan Wang, Bochao Wu, Bei Feng, Chengda Lu, Chenggang Zhao, Chengqi Deng, Chenyu Zhang, Chong Ruan, Damai Dai, Deli Chen, Dongjie Ji, Erhang Li, Fangyun Lin, Fucong Dai, Fuli Luo, Guangbo Hao, Guanting Chen, Guowei Li, H.~Zhang, Han Bao, Hanwei Xu, Haocheng Wang, Honghui Ding, Huajian Xin, Huazuo Gao, Hui Qu, Hui Li, Jianzhong Guo, Jiashi Li, Jiawei Wang, Jingchang Chen, Jingyang Yuan, Junjie Qiu, Junlong Li, J.~L. Cai, Jiaqi Ni, Jian Liang, Jin Chen, Kai Dong, Kai Hu, Kaige Gao, Kang Guan, Kexin Huang, Kuai Yu, Lean Wang, Lecong Zhang, Liang Zhao, Litong Wang, Liyue Zhang, Lei Xu, Leyi Xia, Mingchuan Zhang, Minghua Zhang, Minghui Tang, Meng Li, Miaojun Wang, Mingming Li, Ning Tian, Panpan Huang, Peng Zhang, Qiancheng Wang, Qinyu Chen, Qiushi Du, Ruiqi Ge, Ruisong Zhang,
  Ruizhe Pan, Runji Wang, R.~J. Chen, R.~L. Jin, Ruyi Chen, Shanghao Lu, Shangyan Zhou, Shanhuang Chen, Shengfeng Ye, Shiyu Wang, Shuiping Yu, Shunfeng Zhou, Shuting Pan, S.~S. Li, Shuang Zhou, Shaoqing Wu, Shengfeng Ye, Tao Yun, Tian Pei, Tianyu Sun, T.~Wang, Wangding Zeng, Wanjia Zhao, Wen Liu, Wenfeng Liang, Wenjun Gao, Wenqin Yu, Wentao Zhang, W.~L. Xiao, Wei An, Xiaodong Liu, Xiaohan Wang, Xiaokang Chen, Xiaotao Nie, Xin Cheng, Xin Liu, Xin Xie, Xingchao Liu, Xinyu Yang, Xinyuan Li, Xuecheng Su, Xuheng Lin, X.~Q. Li, Xiangyue Jin, Xiaojin Shen, Xiaosha Chen, Xiaowen Sun, Xiaoxiang Wang, Xinnan Song, Xinyi Zhou, Xianzu Wang, Xinxia Shan, Y.~K. Li, Y.~Q. Wang, Y.~X. Wei, Yang Zhang, Yanhong Xu, Yao Li, Yao Zhao, Yaofeng Sun, Yaohui Wang, Yi~Yu, Yichao Zhang, Yifan Shi, Yiliang Xiong, Ying He, Yishi Piao, Yisong Wang, Yixuan Tan, Yiyang Ma, Yiyuan Liu, Yongqiang Guo, Yuan Ou, Yuduan Wang, Yue Gong, Yuheng Zou, Yujia He, Yunfan Xiong, Yuxiang Luo, Yuxiang You, Yuxuan Liu, Yuyang Zhou, Y.~X. Zhu, Yanhong Xu,
  Yanping Huang, Yaohui Li, Yi~Zheng, Yuchen Zhu, Yunxian Ma, Ying Tang, Yukun Zha, Yuting Yan, Z.~Z. Ren, Zehui Ren, Zhangli Sha, Zhe Fu, Zhean Xu, Zhenda Xie, Zhengyan Zhang, Zhewen Hao, Zhicheng Ma, Zhigang Yan, Zhiyu Wu, Zihui Gu, Zijia Zhu, Zijun Liu, Zilin Li, Ziwei Xie, Ziyang Song, Zizheng Pan, Zhen Huang, Zhipeng Xu, Zhongyu Zhang, and Zhen Zhang. 2025.
\newblock Deepseek-r1: Incentivizing reasoning capability in llms via reinforcement learning.
\newblock \emph{arXiv preprint arXiv:2501.12948}.

\bibitem[{Hang et~al.(2024)Hang, Yu, and Tan}]{10480162}
Ching~Nam Hang, Pei-Duo Yu, and Chee~Wei Tan. 2024.
\newblock Trumorgpt: Query optimization and semantic reasoning over networks for automated fact-checking.
\newblock In \emph{CISS}.

\bibitem[{Hern{\'a}ndez-Casta{\~n}eda et~al.(2017)Hern{\'a}ndez-Casta{\~n}eda, Calvo, Gelbukh, and Flores}]{hernandez2017cross}
{\'A}ngel Hern{\'a}ndez-Casta{\~n}eda, Hiram Calvo, Alexander Gelbukh, and Jorge J~Garc{\'\i}a Flores. 2017.
\newblock Cross-domain deception detection using support vector networks.
\newblock \emph{Soft Computing}.

\bibitem[{Huang and Sun(2023)}]{huang2023harnessing}
Yue Huang and Lichao Sun. 2023.
\newblock Harnessing the power of chatgpt in fake news: An in-depth exploration in generation, detection and explanation.
\newblock \emph{arXiv preprint arXiv:2310.05046}.

\bibitem[{Karisani and Ji(2024)}]{karisani2024fact}
Payam Karisani and Heng Ji. 2024.
\newblock Fact checking beyond training set.
\newblock In \emph{NAACL}.

\bibitem[{Li et~al.(2023)Li, Wang, He, Zhang, and Liu}]{li2023improving}
Jingqiu Li, Lanjun Wang, Jianlin He, Yongdong Zhang, and Anan Liu. 2023.
\newblock Improving rumor detection by class-based adversarial domain adaptation.
\newblock In \emph{ACM MM}.

\bibitem[{Lin et~al.(2022)Lin, Ma, Chen, Yang, Cheng, and Guang}]{lin-etal-2022-detect}
Hongzhan Lin, Jing Ma, Liangliang Chen, Zhiwei Yang, Mingfei Cheng, and Chen Guang. 2022.
\newblock Detect rumors in microblog posts for low-resource domains via adversarial contrastive learning.
\newblock In \emph{NAACL}.

\bibitem[{Liu et~al.(2024{\natexlab{a}})Liu, Wang, Li, and Li}]{liu-etal-2024-teller}
Hui Liu, Wenya Wang, Haoru Li, and Haoliang Li. 2024{\natexlab{a}}.
\newblock {TELLER}: A trustworthy framework for explainable, generalizable and controllable fake news detection.
\newblock In \emph{ACL}.

\bibitem[{Liu et~al.(2024{\natexlab{b}})Liu, Zhu, Zhang, Tang, Zhang, Liu, Liu, and Chen}]{liu2024detect}
Ye~Liu, Jiajun Zhu, Kai Zhang, Haoyu Tang, Yanghai Zhang, Xukai Liu, Qi~Liu, and Enhong Chen. 2024{\natexlab{b}}.
\newblock Detect, investigate, judge and determine: A novel llm-based framework for few-shot fake news detection.
\newblock \emph{arXiv preprint arXiv:2407.08952}.

\bibitem[{Liu et~al.(2024{\natexlab{c}})Liu, Chen, Zhang, Gao, Zhang, and Yan}]{liu2024skepticism}
Yuhan Liu, Xiuying Chen, Xiaoqing Zhang, Xing Gao, Ji~Zhang, and Rui Yan. 2024{\natexlab{c}}.
\newblock From skepticism to acceptance: simulating the attitude dynamics toward fake news.
\newblock In \emph{IJCAI}.

\bibitem[{Liu et~al.(2024{\natexlab{d}})Liu, Song, Zhang, Zhang, Chen, and Yan}]{liu2024stepwise}
Yuhan Liu, Zirui Song, Juntian Zhang, Xiaoqing Zhang, Xiuying Chen, and Rui Yan. 2024{\natexlab{d}}.
\newblock The stepwise deception: Simulating the evolution from true news to fake news with llm agents.
\newblock \emph{arXiv preprint arXiv:2410.19064}.

\bibitem[{Liu et~al.(2024{\natexlab{e}})Liu, Yang, Xie, de~Kock, Ananiadou, and Hovy}]{liu2024raemollm}
Zhiwei Liu, Kailai Yang, Qianqian Xie, Christine de~Kock, Sophia Ananiadou, and Eduard Hovy. 2024{\natexlab{e}}.
\newblock Raemollm: Retrieval augmented llms for cross-domain misinformation detection using in-context learning based on emotional information.
\newblock \emph{arXiv preprint arXiv:2406.11093}.

\bibitem[{Nan et~al.(2021)Nan, Cao, Zhu, Wang, and Li}]{nan2021mdfend}
Qiong Nan, Juan Cao, Yongchun Zhu, Yanyan Wang, and Jintao Li. 2021.
\newblock Mdfend: Multi-domain fake news detection.
\newblock In \emph{CIKM}.

\bibitem[{Nan et~al.(2024)Nan, Sheng, Cao, Hu, Wang, and Li}]{nan2024let}
Qiong Nan, Qiang Sheng, Juan Cao, Beizhe Hu, Danding Wang, and Jintao Li. 2024.
\newblock {Let Silence Speak: Enhancing Fake News Detection with Generated Comments from Large Language Models}.
\newblock In \emph{CIKM}.

\bibitem[{P{\'e}rez-Rosas and Mihalcea(2014)}]{perez2014cross}
Ver{\'o}nica P{\'e}rez-Rosas and Rada Mihalcea. 2014.
\newblock Cross-cultural deception detection.
\newblock In \emph{ACL}.

\bibitem[{Pryzant et~al.(2023)Pryzant, Iter, Li, Lee, Zhu, and Zeng}]{pryzant2023automatic}
Reid Pryzant, Dan Iter, Jerry Li, Yin~Tat Lee, Chenguang Zhu, and Michael Zeng. 2023.
\newblock Automatic prompt optimization with "gradient descent" and beam search.
\newblock \emph{arXiv preprint arXiv:2305.03495}.

\bibitem[{Ran and Jia(2023)}]{ran2023unsupervised}
Hongyan Ran and Caiyan Jia. 2023.
\newblock Unsupervised cross-domain rumor detection with contrastive learning and cross-attention.
\newblock In \emph{AAAI}.

\bibitem[{Ran et~al.(2023)Ran, Jia, and Yu}]{ran2023metric}
Hongyan Ran, Caiyan Jia, and Jian Yu. 2023.
\newblock A metric-learning method for few-shot cross-event rumor detection.
\newblock \emph{Neurocomputing}.

\bibitem[{Su et~al.(2019)Su, Zeng, Xie, Wen, Yin, and Liu}]{su2019exploring}
Jinsong Su, Jiali Zeng, Jun Xie, Huating Wen, Yongjing Yin, and Yang Liu. 2019.
\newblock Exploring discriminative word-level domain contexts for multi-domain neural machine translation.
\newblock \emph{IEEE transactions on pattern analysis and machine intelligence}.

\bibitem[{Wan et~al.(2024)Wan, Feng, Tan, Wang, Tsvetkov, and Luo}]{wan2024dell}
Herun Wan, Shangbin Feng, Zhaoxuan Tan, Heng Wang, Yulia Tsvetkov, and Minnan Luo. 2024.
\newblock {DELL}: Generating reactions and explanations for {LLM}-based misinformation detection.
\newblock In \emph{ACL}.

\bibitem[{Wang et~al.(2025{\natexlab{a}})Wang, Song, Tian, Peng, Yu, Mi, Su, and Yu}]{wang2025litesearch}
Ante Wang, Linfeng Song, Ye~Tian, Baolin Peng, Dian Yu, Haitao Mi, Jinsong Su, and Dong Yu. 2025{\natexlab{a}}.
\newblock Litesearch: Efficient tree search with dynamic exploration budget for math reasoning.
\newblock In \emph{AAAI}.

\bibitem[{Wang et~al.(2025{\natexlab{b}})Wang, Song, Tian, Yu, Mi, Duan, Tu, Su, and Yu}]{wang2025don}
Ante Wang, Linfeng Song, Ye~Tian, Dian Yu, Haitao Mi, Xiangyu Duan, Zhaopeng Tu, Jinsong Su, and Dong Yu. 2025{\natexlab{b}}.
\newblock Don't get lost in the trees: Streamlining llm reasoning by overcoming tree search exploration pitfalls.
\newblock \emph{arXiv preprint arXiv:2502.11183}.

\bibitem[{Wei et~al.(2024)Wei, Yang, Song, Lu, Hu, Huang, Tran, Peng, Liu, Huang et~al.}]{wei2024long}
Jerry Wei, Chengrun Yang, Xinying Song, Yifeng Lu, Nathan Hu, Jie Huang, Dustin Tran, Daiyi Peng, Ruibo Liu, Da~Huang, et~al. 2024.
\newblock Long-form factuality in large language models.
\newblock In \emph{NeurIPS}.

\bibitem[{Wu et~al.(2024)Wu, Guo, and Hooi}]{SheepDog}
Jiaying Wu, Jiafeng Guo, and Bryan Hooi. 2024.
\newblock Fake news in sheep's clothing: Robust fake news detection against llm-empowered style attacks.
\newblock In \emph{ACM SIGKDD}.

\bibitem[{Xu et~al.(2023)Xu, Sun, Zheng, Geng, Zhao, Feng, Tao, and Jiang}]{xu2023wizardlm}
Can Xu, Qingfeng Sun, Kai Zheng, Xiubo Geng, Pu~Zhao, Jiazhan Feng, Chongyang Tao, and Daxin Jiang. 2023.
\newblock Wizardlm: Empowering large language models to follow complex instructions.
\newblock \emph{arXiv preprint arXiv:2304.12244}.

\bibitem[{Yang et~al.(2024)Yang, Zhao, Zhu, Zhou, Xu, Jia, and Zan}]{yang2024zhongjing}
Songhua Yang, Hanjie Zhao, Senbin Zhu, Guangyu Zhou, Hongfei Xu, Yuxiang Jia, and Hongying Zan. 2024.
\newblock Zhongjing: Enhancing the chinese medical capabilities of large language model through expert feedback and real-world multi-turn dialogue.
\newblock In \emph{AAAI}.

\bibitem[{Yue~Huang(2024)}]{poblete2023using}
Lichao~Sun Yue~Huang. 2024.
\newblock Fakegpt: Fake news generation, explanation and detection of large language models.
\newblock In \emph{WWW}.

\bibitem[{Zeng et~al.(2019)Zeng, Liu, Su, Ge, Lu, Yin, and Luo}]{zeng2019iterative}
Jiali Zeng, Yang Liu, Jinsong Su, Yubing Ge, Yaojie Lu, Yongjing Yin, and Jiebo Luo. 2019.
\newblock Iterative dual domain adaptation for neural machine translation.
\newblock In \emph{EMNLP}.

\bibitem[{Zhang and Gao(2023)}]{zhang2023towards}
Xuan Zhang and Wei Gao. 2023.
\newblock Towards llm-based fact verification on news claims with a hierarchical step-by-step prompting method.
\newblock In \emph{IJCNLP}.

\bibitem[{Zhou et~al.(2020)Zhou, Chen, Liu, Xiao, Su, Guo, and Wu}]{zhou2020exploring}
Chulun Zhou, Liang-Yu Chen, Jiachen Liu, Xinyan Xiao, Jinsong Su, Sheng Guo, and Hua Wu. 2020.
\newblock Exploring contextual word-level style relevance for unsupervised style transfer.
\newblock In \emph{ACL}.

\end{thebibliography}

\appendix
\section{{Frequently Asked Questions}}
\label{FAQs}



\subsection{{Why Adopt a Multi-Agent Framework?}}

\begin{table}[htb]
\centering
\small
\begin{tabular}{lcc}
\toprule 
\textbf{} & \textbf{Avg. Task Coverage} &\textbf{F1}  \\
\midrule

Llama-3.1-8B & 0.43 & 46.16 \\
\quad \textit{w/} multi-agent & \textbf{0.92} & \textbf{53.21 }\\
        \arrayrulecolor{gray!70}
        \hdashline[7pt/2pt]
        \arrayrulecolor{black}
Llama-3.1-405B & 0.57 & 76.54 \\
\quad \textit{w/} multi-agent & \textbf{1} & \textbf{80.86} \\
    \arrayrulecolor{gray!70} 
    \hdashline[7pt/2pt]
    \arrayrulecolor{black} GPT-3.5-0125 & 0.53 & 64.08 \\ 
\quad \textit{w/} multi-agent & \textbf{0.99} & \textbf{71.75}\\

\bottomrule
\end{tabular}
\caption{{Comparison of task coverage and F1 between a single LLM and the multi-agent framework.}}
\label{Appendix multi-agent framework}
\end{table}

Misinformation detection involves multi-dimensional analysis of news and the integration of these analyses for judgment. Typically, a single LLM is not capable of handling these complex tasks simultaneously. In contrast, a multi-agent framework decomposes the complex task into simpler subtasks, which are then performed by different expert agents. To verify its effectiveness, we compare the task coverage and detection F1 of the multi-agent framework with a single LLM on Weibo21. For the single LLM, we prompt it to conduct linguistic feature analysis, comment analysis, and fact-checking on the news, and then make a judgment based on the analysis results. As shown in Table \ref{Appendix multi-agent framework}, the task coverage and detection F1 of the multi-agent framework are both significantly higher than those of the single LLM.

\subsection{{How does decision rule optimization differs from transfer learning and domain adaptation approaches?}}

Traditional transfer learning and domain adaptation techniques can be applied to cross-domain misinformation detection. These methods typically improve a model’s generalization ability by updating its parameters. However, when training samples are limited and the model has a large scale of parameters, such traditional approaches are often difficult to apply effectively.

In contrast, our proposed decision rule optimization method is well-suited for this scenario. Instead of updating model parameters, we enhance the generalization ability of Judge Agent in cross-domain misinformation detection by searching for and applying the optimal decision rules.

\subsection{{Efficiency Comparison}}
\label{Efficiency Comparison}

\begin{table}[htb]
\centering
\small
\begin{tabular}{lcc}
\toprule 
\textbf{} & \textbf{Avg. Token} & \textbf{F1} \\
\midrule
RAEmo & 1125 & 52.97 \\
MARO (ours) & \textbf{1047} &\textbf{55.98} \\
\bottomrule
\end{tabular}
\caption{{Efficiency comparison.}}\label{computational_cost}
\end{table}

We conduct a computational cost analysis on Weibo21, comparing MARO with the strongest baseline, RAEmo \cite{liu2024raemollm}. Both methods use Llama-3.1-8B as the underlying model. Specifically, we measure the average number of input tokens required to complete both the training and inference processes, as well as the detection F1. As shown in Table \ref{computational_cost}, compared to RAEmo, MARO reduces the average token consumption by 6.9\% while achieving a 5.7\% improvement in F1.

\section{Prompts}
\label{Appendix A} 

\subsection{System Prompts for the Multi-Dimensional Analysis Module}
\label{Appendix A.1} 

We list the system prompts for the agents in Multi-Dimensional Analysis Module as follows:

\fbox{
\parbox{0.9\linewidth}{\textbf{Linguistic Feature Analysis Agent}

In a multi-agent misinformation detection system, you act as the linguistic feature analysis agent, responsible for conducting an in-depth analysis of the emotional polarity and writing style of the news while generating a linguistic feature analysis report.}
}

\fbox{
\parbox{0.9\linewidth}{\textbf{Comment Analysis Agent}

In a multi-agent misinformation detection system, you act as the comment analysis agent, responsible for conducting an in-depth analysis of commenters' stances and emotional polarity towards the news and identifying fact-checking information within the comments to generate a comment analysis report.}
}

\fbox{
\parbox{0.915\linewidth}{\textbf{Fact-Questioning Agent}

In a multi-agent misinformation detection system, you act as the fact questioning agent, responsible for generating specific yes/no questions based on the statements in the news to assist in determining its authenticity.}}

\fbox{
\parbox{0.9\linewidth}{\textbf{Fact-Checking Agent}

In a multi-agent misinformation detection system, you act as the Fact-Questioning Agent, responsible for analyzing the consistency between statements in news and factual evidence. You need to invoke the Wikipedia tool and leverage clues from the search engine to retrieve relevant facts relevant to the statements. Then, you need assess the consistency between the statements and the facts, producing a fact-checking analysis report.}}

\fbox{
\parbox{0.9\linewidth}{\textbf{Questioning Agent}

In a multi-agent misinformation detection system, you act as the Questioning Agent, responsible for reviewing the source content and the analysis report to identify aspects requiring further investigation. Then, you need to pose targeted questions, encouraging the report providers to perform more in-depth and comprehensive analysis.}}
\begin{table*}[tbp]
    \centering 
 \renewcommand{\arraystretch}{1.2}
\setlength{\tabcolsep}{21pt}
 {\small
\begin{tabular}{lccccc}
\toprule \textbf{Domain} & \textbf{Science} & \textbf{Military} & \textbf{Education} & \textbf{Disasters} & \textbf{Politics} \\
\midrule Real & 143 & 121 & 243 & 185 & 306 \\
Fake & 93 & 222 & 248 & 591 & 546 \\
\midrule All & 236 & 343 & 491 & 776 & 852 \\
\midrule\midrule \textbf{Domain} & \textbf{Health} & \textbf{Finance} & \textbf{Entertain} & \textbf{Society} & \textbf{All} \\
\midrule Real & 485 & 959 & 1000 & 1198 & 4640 \\
Fake & 515 & 362 & 440 & 1471 & 4488 \\
\midrule All & 1000 & 1321 & 1440 & 2669 & 9128 \\
\bottomrule
\end{tabular}}
\caption{Data Statistics of Weibo21.} 
\label{weibo21 dataset}
\end{table*}


\begin{table*}[tbp]
    \centering 
 \renewcommand{\arraystretch}{1.2}
 \setlength{\tabcolsep}{15pt}
  {\small
\begin{tabular}{lccccccccc}
\toprule \textbf{Domain}&\textbf{Tech}&\textbf{Edu}&\textbf{Biz}&\textbf{Sport}&\textbf{Polit}&\textbf{Entmt}&\textbf{Cele}&\textbf{All}\\ \midrule
Legit&40&40&40&40&40&40&250&490\\
Fake&40&40&40&40&40&40&250&490\\ \midrule
All&80&80&80&80&80&80&500&980 \\ 
\bottomrule
\end{tabular}}
\caption{Data Statistics of AMTCele.} 
\label{amt dataset}
\end{table*}
\subsection{Prompt for the Decision Rule Optimization Agent}
\label{Appendix A.2}

\fbox{
\parbox[c]{0.94\linewidth}{ 
\textbf{Decision Rule Optimization Agent}

You have been provided with a set of decision rules and their corresponding accuracy score. The decision rules are ordered by their accuracy in ascending order, where a higher accuracy represents higher generalizability.

\quad

<\textit{decision rule 1, accuracy 1}>

<\textit{decision rule 2, accuracy 2}>

(...more example pairs...)

\quad

Below are several examples demonstrating how to apply these decision rules. In each example, replace <DECISION RULE> with your decision rule, read the input carefully, and generate an accurate judgment. If the judgment matches the provided ground-truth label, it is considered correct; otherwise, it is wrong.

\quad

Input: [example news]

<DECISION RULE> 

Output: fake

(...more examples...)

\quad

Now, design a new decision rule that differs from the existing ones and aim to maximize its accuracy. }
}

\section{Datasets Details}
\label{datasets details}


We conduct experiments on the Weibo21 and AMTCele, respectively. The statistical of both datasets are summarized in Tables \ref{weibo21 dataset} and \ref{amt dataset}.


\section{Baselines}
\label{baselines}

The adopted baselines are listed as follows:

\begin{itemize}
    \item  \noindent\textbf{UCD-RD} \cite{ran2023unsupervised}\quad This method leverages contrastive learning and cross-attention mechanisms to achieve cross-domain rumor detection through feature alignment and domain-invariant feature learning.

    \item  \noindent\textbf{CADA} \cite{li2023improving}\quad It utilizes category alignment and adversarial training to facilitate cross-domain misinformation detection.

    \item  \noindent\textbf{HiSS} \cite{zhang2023towards}\quad Typically, this approach breaks down complex news content into multiple sub-statements and uses search engines to gather clues, progressively verifying each sub-statement to determine the authenticity of the news.


    \item \noindent \textbf{TELLER} \cite{liu-etal-2024-teller}\quad It combines neural-symbolic reasoning with logic rules to enhance explainability and generalizability, providing transparent reasoning paths for misinformation detection.

    \item \noindent \textbf{ADAF} \cite{karisani2024fact}\quad This approach enhances cross-domain fact-checking by adversarially training the retriever for robustness and optimizing the reader to be insensitive to evidence order, improving overall performance across domains.

    \item \noindent \textbf{SAFE} \cite{wei2024long}\quad The model decomposes news content into independent facts and verifies the authenticity of each fact through multi-step reasoning.

    \item \noindent \textbf{DELL} \cite{wan2024dell}\quad It uses LLMs to generate diverse news reactions and interpretable agent tasks, aiming to enhance accuracy and calibration in misinformation detection by integrating expert predictions.

     \item \noindent \textbf{DeepSeek-R1 }\cite{deepseekai2025deepseekr1incentivizingreasoningcapability}\quad  It is a reasoning model that integrates multi-stage training and cold-start data.

    \item \noindent \textbf{RAEmo} \cite{liu2024raemollm}\quad  It constructs a sentiment-embedded retrieval database, leveraging sentiment examples from the source domain for in-context learning to verify content authenticity in the target domain.
\end{itemize}

\section{Cross-Validation Experiments}
\label{Cross-validation Experiments}

\begin{figure}[h!]
    \centering
    \includegraphics[width=0.45\textwidth]{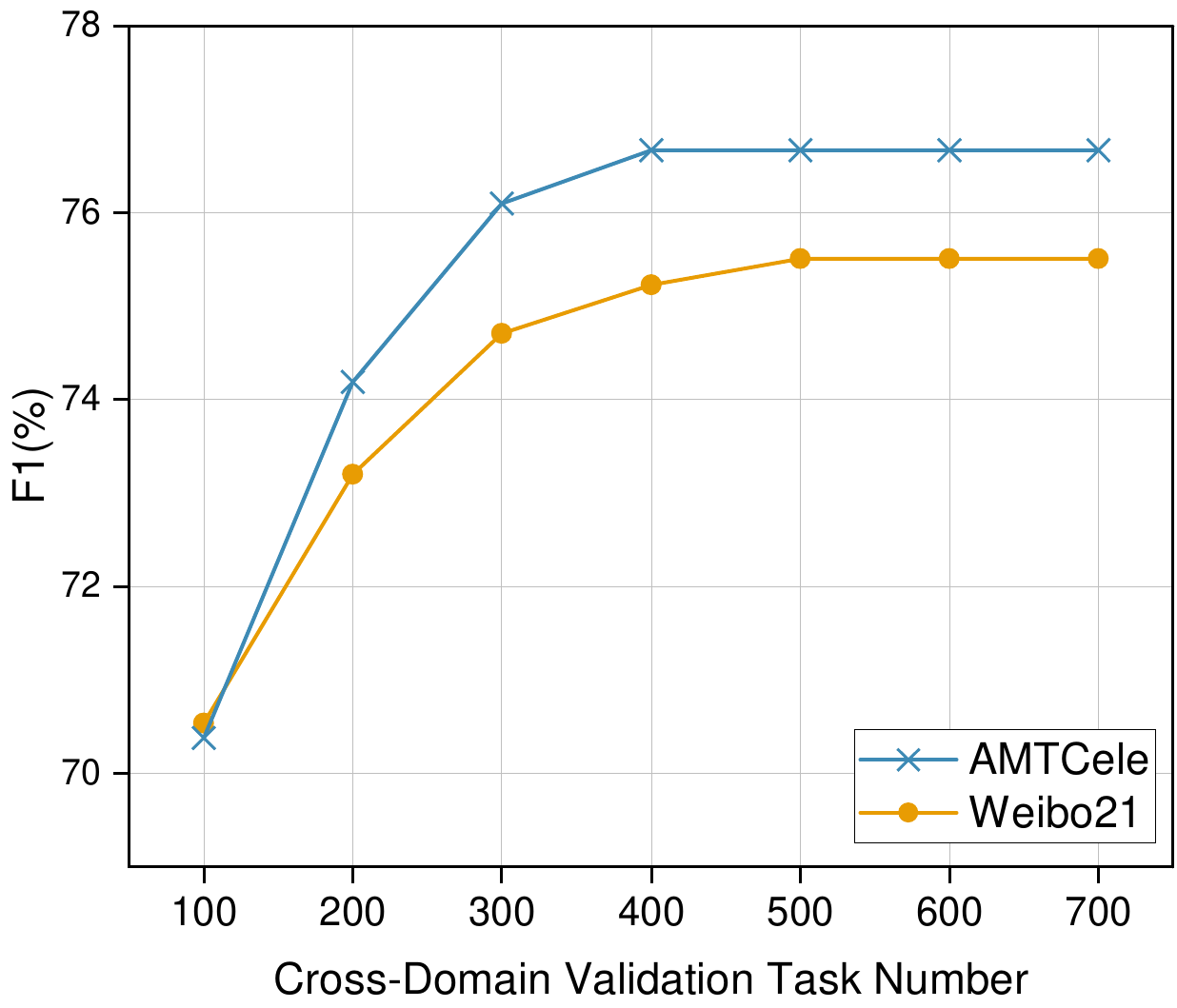}
    \caption{Cross-validation experiments.}
    \label{cross validation}
\end{figure}

To determine the cross-domain validation task number \( N_{vt} \), we conduct 8-fold cross-validation experiments on Weibo21 and 6-fold cross-validation experiments on AMTCele. Through these experiments, we identify \( N_{vt} = 500 \) as the optimal value for Weibo21 and $N_{vt}=400$ for AMTCele, with the validation results illustrated in Figure \ref{cross validation}.

\section{Similarity Matrix}
\label{sim matrix}
\begin{figure}[ht]
    \centering
    \includegraphics[width=0.46\textwidth]{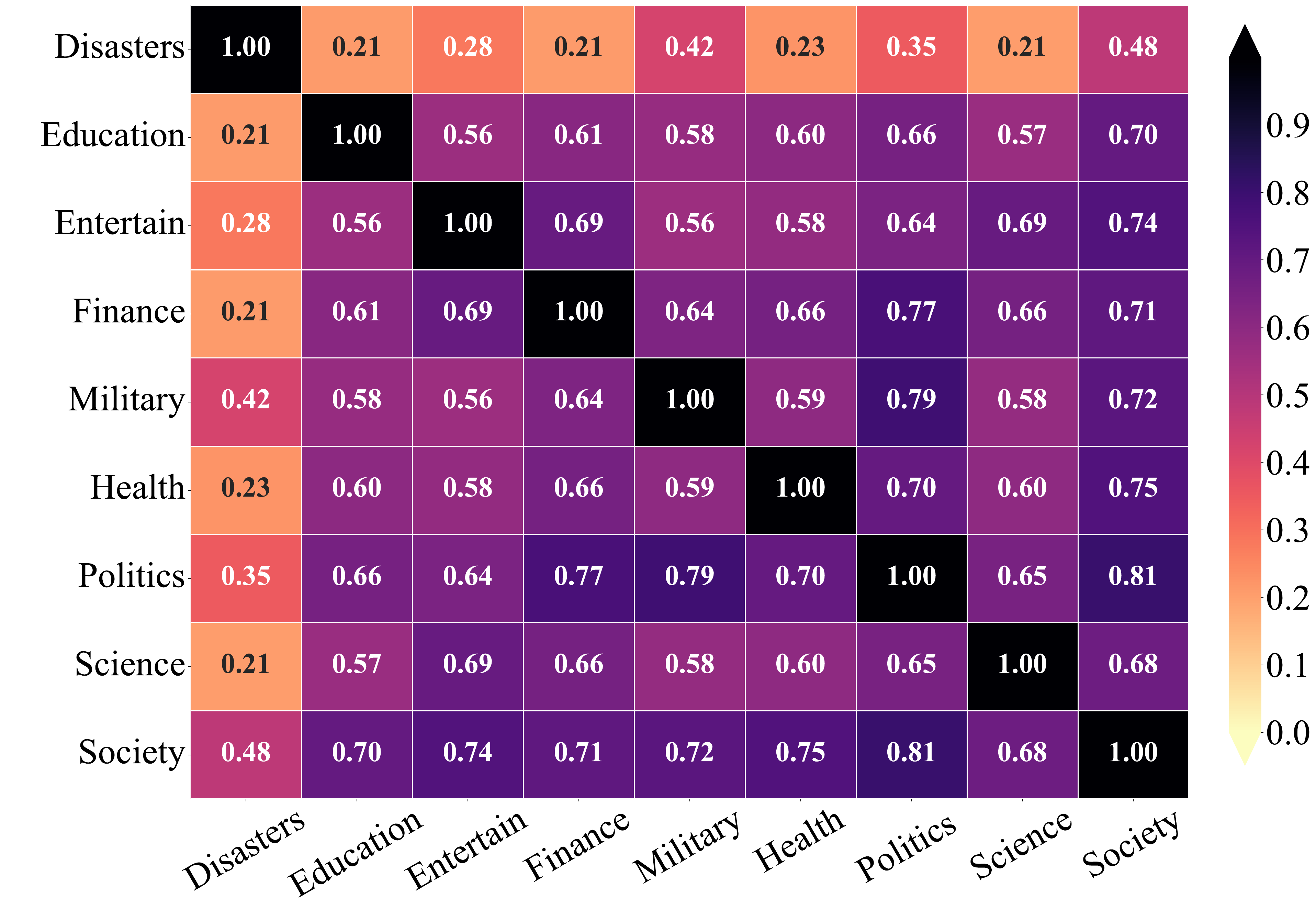}
    \caption{Domain similarity matrix of Weibo21.}
    \label{similarity matrix}
\end{figure}

We compute the domain similarity of the Weibo21 dataset using TF-IDF, with the resulting domain similarity matrix visualized in Figure \ref{similarity matrix}.

\section{More Results}

\subsection{Performance Comparison under Significant Source-Target Domain Differences}
\label{Significant Source-Target Domain Differences}
In order to explore the performance of MARO when there are significant differences between the source and the target domain. According to the domain similarity matrix of Weibo21 (Figure \ref{similarity matrix}), we select Disasters and Education—which are significantly different from all other domains—as the target domains. To avoid experimental redundancy, we choose the three domains with the lowest similarity to each target domain as the source domains (Science, Education, Finance → Disasters; Disaster, Science, Entertain → Education). The experiments are conducted using GPT-3.5-0125 as the underlying model, and the results are shown in Table \ref{10}.
\begin{table}[ht!]
    \centering
        \small
        \tabcolsep 3pt
		\begin{tabular}{ccccc}
			\toprule \multirow{2}{*}{\textbf{Method}} & \multicolumn{2}{c}{\textbf{Sci, Edu, Fin -> Dis}} & \multicolumn{2}{c}{\textbf{Dis, Sci, Ent -> Edu}} \\
			\cmidrule(l){2-3}\cmidrule(l){4-5}
			 & Acc. & F1 & Acc. & F1 \\
			\midrule TELLER & 72.24 & 74.17 & 66.45 & 66.58 \\
			 DELL & 72.26 & 73.05 & {\ul 67.24} & {\ul 67.51} \\
			 RAEmo & {\ul 76.52} & {\ul 77.31} & 65.28 & 66.36 \\
			 \rowcolor{gray!50} MARO (ours) & \textbf{81.86} & \textbf{87.65} & \textbf{74.16} & \textbf{74.83} \\
			\bottomrule
		\end{tabular}
        \caption{Performance Comparison under Significant Source-Target Domain Differences\label{10}}
	\end{table}
\begin{table*}[t!]
    \centering
        \small
		\tabcolsep 3.9pt
		\begin{tabular}{ccccccccccccccc}
			\toprule \multirow{2}{*}{\textbf{Method}} & \multicolumn{2}{c}{\textbf{Dis->Edu}} & \multicolumn{2}{c}{\textbf{Dis->Ent}} & \multicolumn{2}{c}{\textbf{Edu->Dis}} & \multicolumn{2}{c}{\textbf{Edu->Ent}} & \multicolumn{2}{c}{\textbf{Ent->Dis}} & \multicolumn{2}{c}{\textbf{Ent->Edu}} & \multicolumn{2}{c}{\textbf{Avg.}} \\
			\cmidrule(l){2-3}\cmidrule(l){4-5}\cmidrule(l){6-7}\cmidrule(l){8-9}\cmidrule(l){10-11}\cmidrule(l){12-13}\cmidrule(l){14-15}
			 & Acc. & F1 & Acc. & F1 & Acc. & F1 & Acc. & F1 & Acc. & F1 & Acc. & F1 & Acc. & F1 \\
			\midrule TELLER & 65.79 & 65.18 & 59.27 & 59.16 & 73.28 & 73.56 & 59.82 & 59.67 & 73.32 & 73.75 & 65.91 & 65.45 & 66.23 & 66.13 \\
			 DELL & {\ul 69.05} & {\ul 68.31} & {\ul 65.42} & {\ul 64.28} & 74.38 & 73.91 & {\ul 65.52} & {\ul 64.61} & 74.46 & 73.95 & {\ul 69.38} & {\ul 68.57} & {\ul 69.7} & {\ul 68.94} \\
			 RAEmo & 65.11 & 64.39 & 60.24 & 60.87 & {\ul 77.34} & {\ul 76.82} & 61.38 & 61.76 & {\ul 77.95} & {\ul 76.97} & 65.57 & 64.68 & 67.93 & 67.58 \\
			\rowcolor{gray!50} MARO (ours) & \textbf{73.42} & \textbf{74.11} & \textbf{66.24} & \textbf{64.38} & \textbf{81.81} & \textbf{85.82} & \textbf{67.91} & \textbf{66.53} & \textbf{82.16} & \textbf{85.95} & \textbf{73.82} & \textbf{74.25} & \textbf{74.23} & \textbf{75.17} \\
			\bottomrule
		\end{tabular}
        \caption{Performance comparison between MARO and baselines in single-source, single-target domain scenarios\label{single-source}}
	\end{table*}
    
\begin{table*}[t!]
\centering
\small
\setlength\tabcolsep{4.8pt}
\begin{tabular}{l|cccccccccccccc}
\toprule
\multicolumn{1}{c|}{\multirow{2}{*}{\textbf{Method}}} & \multicolumn{2}{c}{\textbf{Disasters}}& & \multicolumn{2}{c}{\textbf{Entertain}}& & \multicolumn{2}{c}{\textbf{Health}} &  & \multicolumn{2}{c}{\textbf{Politics}} && \multicolumn{2}{c}{\textbf{Society}} \\ \cmidrule(l){2-3}  \cmidrule(l){5-6}   \cmidrule(l){8-9}  \cmidrule(l){11-12}   \cmidrule(l){14-15} 
                                  & \textbf{Acc.}       & \textbf{F1}    &   & \textbf{Acc.}         & \textbf{F1}    &      & \textbf{Acc.}      & \textbf{F1}  &     & \textbf{Acc.}      & \textbf{F1}    &   & \textbf{Acc.}      & \textbf{F1}      \\ \midrule

TELLER \cite{liu-etal-2024-teller}                            & 78.29              & 78.05      &       &  {\ul 83.26}                & \textbf {83.28}     &    & 80.4             & 80.58      &       & 81.26             & 80.46      &       & 76.59             & 77.32            \\
DELL \cite{wan2024dell}                              & 81.05              & 81.26   &          & 82.06                & 81.89    &           & {\ul 83.1}       & {\ul 82.97}    &   & 84.18             & 83.89         &    & \textbf {78.56}             & 77.19            \\
RAEmo \cite{liu2024raemollm}                          & {\ul 84.26}        & {\ul 84.49}   &    & 83.11                & {\ul 82.84}    &           & 83             & 82.85     &        & {\ul 84.75}       & {\ul 84.55}  &     & 77.65             & {\ul 77.49}            \\
\rowcolor{gray!50}
MARO (ours)                              & \textbf{85.05}     & \textbf{89.66} &   & \textbf{86.53}       & {79.7}   &   & \textbf{86.8}    & \textbf{87.15}  &  & \textbf{88.01}    & \textbf{90.41} &  & {\ul 77.97}    & \textbf{79.61}   \\ \midrule \midrule
\multicolumn{1}{c|}{\multirow{2}{*}{\textbf{Method}}} & \multicolumn{2}{c}{\textbf{Education}} && \multicolumn{2}{c}{\textbf{Finance}}   &    & \multicolumn{2}{c}{\textbf{Military}}& & \multicolumn{2}{c}{\textbf{Science}}&  & \multicolumn{2}{c}{\textbf{Avg.}}    \\ \cmidrule(l){2-3}  \cmidrule(l){5-6}   \cmidrule(l){8-9}  \cmidrule(l){11-12}   \cmidrule(l){14-15} 
                                  & \textbf{Acc.}       & \textbf{F1}  &     & \textbf{Acc.}         & \textbf{F1}  &       & \textbf{Acc.}      & \textbf{F1}   &    & \textbf{Acc.}      & \textbf{F1}    &   & \textbf{Acc.}      & \textbf{F1}      \\ \midrule

TELLER \cite{liu-etal-2024-teller}                            & 72.11              & 72.12     &        & 74.76                & 75.25       &        & 88.19             & 87.89    &         & 72.21             & 71.19       &      & 78.56             & 78.46            \\
 DELL \cite{wan2024dell}                              & 74.64              & 75.12       &      & \textbf{80.91}       & {\ul 79.24}   &            & 87.98             & 88.26      &       & 71.88             & \textbf{72.58}       &      & 80.48             & 80.27            \\
RAEmo \cite{liu2024raemollm}                          & {\ul 75.46}        & {\ul 75.88}  &     & {\ul 80.35}                & \textbf{79.75}  &       & {\ul 88.94}          & {\ul 89.05}     & &\textbf {72.75}            & {\ul 71.75}      &       & {\ul 81.14}       & {\ul 80.96}      \\
\rowcolor{gray!50}
MARO (ours)                              & \textbf{76.82}     & \textbf{77.1}  &  & {77.76}          & {65.89}   &   & \textbf{91.44}    & \textbf{93.39} &   & {\ul 72.57}    & {67.66}  &  & \textbf{82.56}    & \textbf{81.18}   \\ \bottomrule
\end{tabular}
\caption{Performance comparison on Weibo21 using LLaMA-3.1-405B as the underlying model.}
\label{table Weibo21 405B}
\end{table*}
\begin{table*}[t!]
\centering
\small
\setlength\tabcolsep{7.5pt}
\begin{tabular}{l|ccccccccccc}
\toprule
\multicolumn{1}{c|}{\multirow{2}{*}{\textbf{Method}}} & \multicolumn{2}{c}{\textbf{Biz}}&   & \multicolumn{2}{c}{\textbf{Edu}}  & & \multicolumn{2}{c}{\textbf{Cele}} && \multicolumn{2}{c}{\textbf{Entmt}} \\  \cmidrule(l){2-3}  \cmidrule(l){5-6}   \cmidrule(l){8-9}  \cmidrule(l){11-12} 
                              & \textbf{Acc.}     & \textbf{F1}   &  & \textbf{Acc.}     & \textbf{F1}  &   & \textbf{Acc.}    & \textbf{F1}  &   & \textbf{Acc.}     & \textbf{F1}     \\ \midrule

TELLER \cite{liu-etal-2024-teller}                        & {\ul 83.15}      & 83.19       &    & 82.75            & 82.81    &       & 77.6           & 77.12     &      & 77.51            & 77.57           \\
DELL \cite{wan2024dell}                          & 83.07            & {\ul 83.75}   &  & {\ul 84.18}            & {\ul 84.35}      &     & 77.2           & 76.82     &      & {\ul 77.94}      & {\ul 78.06}     \\ 
RAEmo \cite{liu2024raemollm}                      & 82.51            & 82.53      &     & 84.09            & 84.15    &       & {\ul 79.4}     & {\ul 79.11}  &   & 75.15            & 75.26           \\ 
\rowcolor{gray!50}
MARO (ours)                          & \textbf{86.25}   & \textbf{86.54} & & \textbf{86.25}   & \textbf{86.11}&  & \textbf{81.2}  & \textbf{80.84} & & \textbf{78.75}   & \textbf{78.81}  \\ \midrule \midrule
\multicolumn{1}{c|}{\multirow{2}{*}{\textbf{Method}}} & \multicolumn{2}{c}{\textbf{Polit}}& & \multicolumn{2}{c}{\textbf{Sport}} && \multicolumn{2}{c}{\textbf{Tech}}& & \multicolumn{2}{c}{\textbf{Avg.}}  \\  \cmidrule(l){2-3}  \cmidrule(l){5-6}   \cmidrule(l){8-9}  \cmidrule(l){11-12}  
                              & \textbf{Acc.}     & \textbf{F1}  &   & \textbf{Acc.}     & \textbf{F1}  &   & \textbf{Acc.}    & \textbf{F1}  &   & \textbf{Acc.}     & \textbf{F1}     \\ \midrule

TELLER \cite{liu-etal-2024-teller}                        & 83.69            & {\ul 86.72} &    & 82.75            & 82.89       &    & 85.57           & 85.56     &      & 81.79            & 82.27           \\
DELL \cite{wan2024dell}                          & 84.06            & 83.91        &   & 84.13            & 84.24    &       & 86.41           & 86.32      &     & 82.43            & 82.49           \\
RAEmo \cite{liu2024raemollm}                      & {\ul 84.31}      & 84.34       &    & {\ul 85.11}      & {\ul 85.16}   &        & {\ul 87.02}           & {\ul 86.91}    & & {\ul 82.47}      & {\ul 82.49}           \\
\rowcolor{gray!50}
MARO (ours)                          & \textbf{87.52}   & \textbf{87.51}  && \textbf{87.35}   & \textbf{87.42}  && \textbf{91.25}  & \textbf{90.87} & & \textbf{85.49}   & \textbf{85.44}  \\ \bottomrule
\end{tabular}
\caption{Performance comparison on AMTCele using LLaMA-3.1-405B as the underlying model.}
\label{table amt 405B}
\end{table*}
    \begin{table*}[t!]
\label{tab3}
\centering
\small
\setlength\tabcolsep{4.8pt}
\begin{tabular}{l|ccccccccccccccccc}
\toprule
\multicolumn{1}{c|}{\multirow{2}{*}{\textbf{Method}}} & \multicolumn{2}{c}{\textbf{Disasters}} & &\multicolumn{2}{c}{\textbf{Entertain}} && \multicolumn{2}{c}{\textbf{Health}} && \multicolumn{2}{c}{\textbf{Politics}} && \multicolumn{2}{c}{\textbf{Society}} \\
\cmidrule(l){2-3}  \cmidrule(l){5-6}   \cmidrule(l){8-9}  \cmidrule(l){11-12} \cmidrule(l){14-15}
& \textbf{Acc.} & \textbf{F1} && \textbf{Acc.} & \textbf{F1} && \textbf{Acc.} & \textbf{F1} && \textbf{Acc.} & \textbf{F1} && \textbf{Acc.} & \textbf{F1} \\\midrule
 TELLER \cite{liu-etal-2024-teller}  & 85.47 & 85.16 & &84.19 & 84.26 && 84.8 & 84.26 && 77.65 & 77.42 && 73.94 & 73.81 \\
 DELL \cite{wan2024dell}  & 86.51 & 86.19 && 86.93 & \textbf{86.89} && {\ul 86.4} & {\ul 86.27} && 79.29 & 79.34 && 75.27 & 75.19 \\
 RAEmo \cite{liu2024raemollm}  & {\ul 87.65} & {\ul 86.27} && \textbf{87.49} & {\ul 85.06} && \textbf{87.2} & \textbf{87.45} && {\ul80.85} & {\ul81.06} && {\ul76.26} & {\ul75.84} \\
\rowcolor{gray!50}  MARO (ours) & \textbf{89.17} & \textbf{92.75} && {\ul 87.43} & 82.5 && {86.2} & {79.08} && \textbf{87.61} & \textbf{91.32} && \textbf{78.48} & \textbf{78.29} \\ \midrule  \midrule
\multicolumn{1}{c|}{\multirow{2}{*}{\textbf{Method}}} & \multicolumn{2}{c}{\textbf{Education}} && \multicolumn{2}{c}{\textbf{Finance}} && \multicolumn{2}{c}{\textbf{Military}} && \multicolumn{2}{c}{\textbf{Science}} && \multicolumn{2}{c}{\textbf{Avg.}} \\
\cmidrule(l){2-3}  \cmidrule(l){5-6}   \cmidrule(l){8-9}  \cmidrule(l){11-12} \cmidrule(l){14-15}    & \textbf{Acc.} & \textbf{F1} && \textbf{Acc.} & \textbf{F1} && \textbf{Acc.} & \textbf{F1} && \textbf{Acc.} & \textbf{F1} && \textbf{Acc.} & \textbf{F1} \\ \midrule
 TELLER \cite{liu-etal-2024-teller}  & 79.55 & 78.94 && 82.97 & 82.84 && 83.19 & 83.13 && 70.38 & 70.54 && 80.24 & 80.04 \\
 DELL \cite{wan2024dell}  & \textbf{81.95} & {\ul81.78} && 83.46 & {\ul 83.57} && 83.49 & 83.25 && 71.15 & 71.05 && 81.61 & 81.5 \\
 RAEmo \cite{liu2024raemollm}  & 81.17 & 81.36 && {\ul84.39} & \textbf{84.16} && {\ul85.51} & {\ul85.34} && {\ul72.78} & {\ul72.54} && {\ul82.58} & {\ul82.12} \\
\rowcolor{gray!50} MARO (ours) & {\ul 81.7} & \textbf{82.62} && \textbf{86.15} & {79.08} && \textbf{87.61} & \textbf{91.32} && \textbf{78.48} & \textbf{78.29} && \textbf{85.03} & \textbf{85.23} \\
\bottomrule
\end{tabular}
\caption{Performance comparison on Weibo21 using Claude-3.5-Sonnet as the underlying model.}
\label{table weibo21 Claude-3.5}
\end{table*}
\begin{table*}[t!]
\label{tab4}
\centering 
\small
\setlength\tabcolsep{7.5pt}
\begin{tabular}{l|ccccccccccc}
\toprule
\multicolumn{1}{c|}{\multirow{2}{*}{\textbf{Method}}} & \multicolumn{2}{c}{\textbf{Biz}} && \multicolumn{2}{c}{\textbf{Edu}} && \multicolumn{2}{c}{\textbf{Cele}} && \multicolumn{2}{c}{\textbf{Entmt}} \\
\cmidrule(l){2-3}  \cmidrule(l){5-6}   \cmidrule(l){8-9}  \cmidrule(l){11-12}   
& \textbf{Acc.} & \textbf{F1} && \textbf{Acc.} & \textbf{F1} && \textbf{Acc.} & \textbf{F1} && \textbf{Acc.} & \textbf{F1}  \\ \midrule
 TELLER \cite{liu-etal-2024-teller}  & 86.55 & 86.27 && 88.54 & 88.69 && 73.2 & 73.51 && 79.28 & 79.62  \\
 DELL \cite{wan2024dell}  & 85.43 & 85.37 && 89.28 & 89.75 && 75.4 & 75.19 && 78.51 & 78.29  \\
 RAEmo \cite{liu2024raemollm}  & {\ul88.25} & {\ul88.41} && {\ul91.38} & {\ul91.05} && {\ul76.2} & {\ul76.38} && {\ul81.79} & {\ul81.64} \\
\rowcolor{gray!50}  MARO (ours) & \textbf{91.68} & \textbf{91.79} && \textbf{92.51} & \textbf{92.31} && \textbf{79.6} & \textbf{79.25} && \textbf{85.25} & \textbf{84.93} \\
 \midrule  \midrule
\multicolumn{1}{c|}{\multirow{2}{*}{\textbf{Method}}} & \multicolumn{2}{c}{\textbf{Polit}}& & \multicolumn{2}{c}{\textbf{Sport}} && \multicolumn{2}{c}{\textbf{Tech}} && \multicolumn{2}{c}{\textbf{Avg.}} \\ 
\cmidrule(l){2-3}  \cmidrule(l){5-6}   \cmidrule(l){8-9}  \cmidrule(l){11-12}
& \textbf{Acc.} & \textbf{F1} && \textbf{Acc.} & \textbf{F1} && \textbf{Acc.} & \textbf{F1} && \textbf{Acc.} & \textbf{F1}  \\ \midrule
TELLER \cite{liu-etal-2024-teller}  & 80.26 & 80.11 && 84.51 & 84.63 && 85.59 & 85.26 && 82.56 & 82.58\\
DELL \cite{wan2024dell}  & 81.57 & 81.95 && 86.26 & 86.39 && 86.47 & 86.42 && 83.27 & 83.34\\
RAEmo \cite{liu2024raemollm}  & {\ul84.35} & {\ul85.16} && {\ul88.26} & {\ul88.81} && {\ul88.54} & {\ul89.06} && {\ul85.54} & {\ul85.79} \\
\rowcolor{gray!50}  MARO (ours) & \textbf{87.65} & \textbf{87.29} && \textbf{89.67} & \textbf{89.75} && \textbf{91.69} & \textbf{91.81} && \textbf{88.29} & \textbf{88.16} \\ \bottomrule
\end{tabular}
\caption{Performance comparison on AMTCele using Claude-3.5-Sonnet as the underlying model.}
\label{table amt Claude-3.5}
\end{table*}
\begin{table*}[ht!]
\label{tab5}
\centering 
\small
\setlength\tabcolsep{4.8pt}
\begin{tabular}{l|ccccccccccccccccc}
\toprule 
\multicolumn{1}{c|}{\multirow{2}{*}{\textbf{Method}}} & \multicolumn{2}{c}{\textbf{Disasters}} && \multicolumn{2}{c}{\textbf{Entertain}} && \multicolumn{2}{c}{\textbf{Health}} && \multicolumn{2}{c}{\textbf{Politics}} && \multicolumn{2}{c}{\textbf{Society}} \\
\cmidrule(l){2-3}  \cmidrule(l){5-6}   \cmidrule(l){8-9}  \cmidrule(l){11-12} \cmidrule(l){14-15}
& \textbf{Acc.} & \textbf{F1} &&\textbf{Acc.} & \textbf{F1} & & \textbf{Acc.} & \textbf{F1} & & \textbf{Acc.} & \textbf{F1} & & \textbf{Acc.} & \textbf{F1}   \\ \midrule
TELLER \cite{liu-etal-2024-teller}  & 52.98 & 53.31 && 51.26 & {\ul 51.54} && 56.4 & 56.51 && 59.52 & 59.46 && 41.05 & 41.18  \\
 DELL \cite{wan2024dell}  & {\ul53.89} & 53.76 && {\ul52.54} & \textbf{52.63} && 55.4 & 55.21 && 61.55 & 61.48 && {\ul 48.24} & {\ul 48.51}  \\
 RAEmo \cite{liu2024raemollm}  & 53.28 & {\ul53.95} && 51.39 & 51.44 && {\ul58.5} & {\ul58.33} && {\ul62.54} & {\ul62.69} && 43.56 & 43.19 \\
\rowcolor{gray!50}  MARO (ours) & \textbf{60.05} & \textbf{69.6} && \textbf{59.73} & {47.52} && \textbf{67.8} & \textbf{68.49} && \textbf{66.03} & \textbf{71.41} && \textbf{49.96} & \textbf{52.42} \\
 \midrule \midrule
\multicolumn{1}{c|}{\multirow{2}{*}{\textbf{Method}}} & \multicolumn{2}{c}{\textbf{Education}}& & \multicolumn{2}{c}{\textbf{Finance}} && \multicolumn{2}{c}{\textbf{Military}} && \multicolumn{2}{c}{\textbf{Science}} && \multicolumn{2}{c}{\textbf{Avg.}} \\
\cmidrule(l){2-3}  \cmidrule(l){5-6}   \cmidrule(l){8-9}  \cmidrule(l){11-12} \cmidrule(l){14-15}
& \textbf{Acc.} & \textbf{F1}& &\textbf{Acc.} & \textbf{F1} & & \textbf{Acc.} & \textbf{F1} & & \textbf{Acc.} & \textbf{F1} & & \textbf{Acc.} & \textbf{F1}   \\ \midrule
TELLER \cite{liu-etal-2024-teller}  & 39.26 & 39.28 && 59.53 & 59.64 && 49.37 & 49.82 && 52.74 & 52.68 && 51.35 & 51.49\\
DELL \cite{wan2024dell}  & {\ul 41.24} & {\ul 41.39} && {\ul 60.25} & {\ul 60.17} && {\ul52.51} & {\ul52.55} && \textbf{53.69} & \textbf{53.57} && {\ul53.25} & {\ul53.26}\\
RAEmo \cite{liu2024raemollm}  & 40.92 & 40.79 && \textbf{61.46} & \textbf{62.28} && 50.52 & 50.73 && {\ul 53.25} &{\ul  53.37} && 52.9 & 52.97 \\
\rowcolor{gray!50}  MARO (ours) & \textbf{46.74} & \textbf{46.96} && {55.14} & {40.4} && \textbf{53.98} & \textbf{60.6} && {52.32} & {46.44} && \textbf{56.86} & \textbf{55.98 }\\ 
\bottomrule
\end{tabular}
\caption{Performance comparison on Weibo21 using LLaMA-3.1-8B as the underlying model.}
\label{table weibo21 8B}
\end{table*}
\begin{table*}[t!]
\label{tab6}
\centering 
\small
\setlength\tabcolsep{7.5pt}
\begin{tabular}{l|ccccccccccc}
\toprule
\multicolumn{1}{c|}{\multirow{2}{*}{\textbf{Method}}}  & \multicolumn{2}{c}{\textbf{Biz}} && \multicolumn{2}{c}{\textbf{Edu}}& & \multicolumn{2}{c}{\textbf{Cele}}& & \multicolumn{2}{c}{\textbf{Entmt}} \\
\cmidrule(l){2-3}  \cmidrule(l){5-6}   \cmidrule(l){8-9}  \cmidrule(l){11-12}   
& \textbf{Acc.} & \textbf{F1} && \textbf{Acc.} & \textbf{F1} && \textbf{Acc.} & \textbf{F1} && \textbf{Acc.} & \textbf{F1}  \\ \midrule
 TELLER \cite{liu-etal-2024-teller}  & 51.29 & 51.42 && 54.52 & 54.39& & 71.4 & 71.42 && 50.69 & 50.73  \\
 DELL \cite{wan2024dell}  & {\ul54.96} & {\ul54.85} && {\ul55.21} & {\ul55.13} && {\ul72.4} & {\ul72.19} && 52.54 & 52.43 \\
 RAEmo \cite{liu2024raemollm}  & 51.63 & 50.86 && 53.22 & 52.59 && 70.8 & 70.34 && {\ul53.24} & {\ul53.61} \\
\rowcolor{gray!50}  MARO (ours) & \textbf{56.79} & \textbf{56.41} && \textbf{57.05} & \textbf{56.38} && \textbf{76.2} & \textbf{75.7} && \textbf{57.95} & \textbf{57.31}  \\
 \midrule \midrule
\multicolumn{1}{c|}{\multirow{2}{*}{\textbf{Method}}} & \multicolumn{2}{c}{\textbf{Polit}} && \multicolumn{2}{c}{\textbf{Sport}}& & \multicolumn{2}{c}{\textbf{Tech}}& & \multicolumn{2}{c}{\textbf{Avg.}} \\  
\cmidrule(l){2-3}  \cmidrule(l){5-6}   \cmidrule(l){8-9}  \cmidrule(l){11-12}   
& \textbf{Acc.} & \textbf{F1} && \textbf{Acc.} & \textbf{F1} && \textbf{Acc.} & \textbf{F1} && \textbf{Acc.} & \textbf{F1}  \\ \midrule
 TELLER \cite{liu-etal-2024-teller}  & 49.56 & 49.79 && 57.54 & 57.36 && 63.27 & 63.18 && 56.9 & 56.9\\
 DELL \cite{wan2024dell}  & {\ul50.94} & {\ul50.78} && 58.25 & 58.37 && {\ul66.74} & {\ul66.28} && {\ul58.72} & {\ul58.58} \\
 RAEmo \cite{liu2024raemollm}  & 50.14 & 50.27 && {\ul60.85} & {\ul60.92} && 61.89 & 61.42 && 57.4 & 57.14 \\
\rowcolor{gray!50} MARO (ours) & \textbf{56.17} & \textbf{55.05} && \textbf{63.29} & \textbf{63.51} && \textbf{70.26} & \textbf{69.65} && \textbf{62.53} & \textbf{62}\\ 
 \bottomrule
\end{tabular}
\caption{Performance comparison on AMTCele using LLaMA-3.1-8B as the underlying model.}
\label{table wamt 8B}
\end{table*}
\subsection{Performance Comparison under single-source, single-target domain}
To further explore performance in single-source and target domain scenarios, we randomly select three domains (Disasters, Education, and Entertainment) from Weibo21, forming six source-target domain pairs. We use GPT-3.5-0125 as the underlying model. The experimental results are shown in Table \ref{single-source}.

\begin{table*}[t!]
    \centering 
\small
\renewcommand{\arraystretch}{1.05}
\setlength{\tabcolsep}{7pt}
\begin{tabular}{lcccccc}
\toprule \textbf{Events}&\textbf{Charlie Hebdo}&\textbf{Sydney Siege}&\textbf{Ferguson}&\textbf{Ottawa Shooting}&\textbf{Germanwings Crash} &\textbf{All}\\ \midrule
Rumors&458&522&284&470&238&1972\\
Non-rumors&1621&699&859&420&231&3830\\ \midrule
All&2079&1221&1143&890&469&5802\\
\bottomrule
\end{tabular}
\caption{Data Statistics of PHEME.} 
\label{pheme dataset}
\end{table*}

\begin{table*}[ht!]
\centering
\small
\setlength{\tabcolsep}{11.5pt}
\renewcommand{\arraystretch}{1.05}
\begin{tabular}{l|cccccccc}
\toprule
\multicolumn{1}{c|}{\multirow{2}{*}{\textbf{Method}}}  & \multicolumn{2}{c}{\textbf{Charlie Hebdo}} &&  \multicolumn{2}{c}{\textbf{Ferguson}}  && \multicolumn{2}{c}{\textbf{Germanwings Crash }} \\
\cmidrule(l){2-3} \cmidrule(l){5-6} \cmidrule(l){8-9}
& \textbf{Acc.} & \textbf{F1}  &&\textbf{Acc.} & \textbf{F1}  && \textbf{Acc.} & \textbf{F1} \\
\midrule
TELLER  \cite{liu-etal-2024-teller}  & 63.17&	62.89	&&61.56	& 60.29&&	58.36	&57.82\\
DELL   \cite{wan2024dell}    &  63.68&	63.05	&&62.18&	61.26	&&59.76	&58.82\\
RAEmo   \cite{liu2024raemollm}   & {\ul 64.36}&	{\ul 63.79}&&	{\ul 63.76} &	{\ul 62.87} &&	{\ul 61.79} &	{\ul 60.68} \\
\rowcolor{gray!50} 
MARO (ours) & \textbf{66.12}&	\textbf{64.86} &&	\textbf{65.11}&	\textbf{64.63} &&	\textbf{63.26} &\textbf{62.83} \\
\midrule
\midrule
\multicolumn{1}{c|}{\multirow{2}{*}{\textbf{Method}}}  & \multicolumn{2}{c}{\textbf{Ottawa Shooting}} &&  \multicolumn{2}{c}{\textbf{Sydney Siege}}  && \multicolumn{2}{c}{\textbf{Avg.}} \\
\cmidrule(l){2-3} \cmidrule(l){5-6} \cmidrule(l){8-9}
& \textbf{Acc.} & \textbf{F1}  &&\textbf{Acc.} & \textbf{F1}  && \textbf{Acc.} & \textbf{F1} \\
\midrule
TELLER  \cite{liu-etal-2024-teller}   & 58.05	& 57.92&&	60.38 &	59.26	&&60.3	&59.64\\
DELL \cite{wan2024dell}      & 59.26&	57.08	&& 60.24	& 58.29&&	61.02&	59.7  \\
RAEmo \cite{liu2024raemollm}     & {\ul 59.56}&	{\ul 58.32}&&	{\ul 61.34} &	{\ul 60.89}&&	{\ul 62.16}&	{\ul 61.31} \\
\rowcolor{gray!50} 
MARO (ours) & \textbf{61.39}&	\textbf{61.28}&&	\textbf{62.76}&	\textbf{61.62}&&	\textbf{63.73}&	\textbf{63.04} \\
\bottomrule
\end{tabular}
\caption{{Performance comparison on PHEME using GPT-3.5-turbo-0125 as the
underlying model.}}
\label{pheme_gpt3.5}
\end{table*}
\begin{table*}[ht!]
\centering
\small
\setlength{\tabcolsep}{11.5pt}
\renewcommand{\arraystretch}{1.05}
\begin{tabular}{l|cccccccc}
\toprule
\multicolumn{1}{c|}{\multirow{2}{*}{\textbf{Method}}}  & \multicolumn{2}{c}{\textbf{Charlie Hebdo}} &&  \multicolumn{2}{c}{\textbf{Ferguson}}  && \multicolumn{2}{c}{\textbf{Germanwings Crash }} \\
\cmidrule(l){2-3} \cmidrule(l){5-6} \cmidrule(l){8-9}
& \textbf{Acc.} & \textbf{F1}  &&\textbf{Acc.} & \textbf{F1}  && \textbf{Acc.} & \textbf{F1} \\
\midrule
TELLER  \cite{liu-etal-2024-teller}  & 63.79	&63.26&&	63.59	&62.68	&&61.37&	60.59\\
DELL   \cite{wan2024dell}    & {\ul 64.38}&	63.75&&	64.05	& 62.97	&&  62.66	& 61.89 \\
RAEmo   \cite{liu2024raemollm}    & 64.27&	{\ul 63.85}&&	{\ul 64.51} &	\textbf{63.26} &&	{\ul 62.79} &	{\ul 62.11} \\
\rowcolor{gray!50} MARO (ours) & \textbf{66.56}&	\textbf{64.86} && 	\textbf{64.63} & 	{\ul 63.03} &&	\textbf{63.26} &	\textbf{63.49} \\
\midrule \midrule
\multicolumn{1}{c|}{\multirow{2}{*}{\textbf{Method}}}  & \multicolumn{2}{c}{\textbf{Ottawa Shooting}} && \multicolumn{2}{c}{\textbf{Sydney Siege}} && \multicolumn{2}{c}{\textbf{Avg.}} \\
\cmidrule(l){2-3} \cmidrule(l){5-6} \cmidrule(l){8-9}
& \textbf{Acc.} & \textbf{F1}  &&\textbf{Acc.} & \textbf{F1}  && \textbf{Acc.} & \textbf{F1} \\
\midrule
TELLER  \cite{liu-etal-2024-teller}  & 59.52	& 59.48&&	60.18	&60.05	&& 61.69	&61.21\\
DELL   \cite{wan2024dell}    & 60.32&	59.56&&	60.27&	59.38 &&	62.34	& 61.51 \\
RAEmo   \cite{liu2024raemollm}    & {\ul 61.29} &	{\ul 59.78}	&& {\ul 60.75}&	{\ul 60.08}	&&{\ul 62.72}&	{\ul 61.82}\\
\rowcolor{gray!50} MARO (ours) & \textbf{61.39} &	\textbf{60.06}&&	\textbf{62.76}&	\textbf{61.62}&	& \textbf{63.72}&	\textbf{62.61} \\
\bottomrule
\end{tabular}
\caption{{Performance comparison on PHEME using Claude-3.5-Sonnet as the
underlying model.}}
\label{pheme_Claude3.5}
\end{table*}
\begin{table*}[ht!]
\centering
\small
\setlength{\tabcolsep}{11.5pt}
\renewcommand{\arraystretch}{1.05}
\begin{tabular}{l|cccccccc}
\toprule
\multicolumn{1}{c|}{\multirow{2}{*}{\textbf{Method}}}  & \multicolumn{2}{c}{\textbf{Charlie Hebdo}} &&  \multicolumn{2}{c}{\textbf{Ferguson}}  && \multicolumn{2}{c}{\textbf{Germanwings Crash }} \\
\cmidrule(l){2-3} \cmidrule(l){5-6} \cmidrule(l){8-9}
& \textbf{Acc.} & \textbf{F1}  &&\textbf{Acc.} & \textbf{F1}  && \textbf{Acc.} & \textbf{F1} \\
\midrule
TELLER  \cite{liu-etal-2024-teller}   & 64.26	& {\ul 63.89}&&	63.72	& 62.91&&	60.57	&60.08 \\
DELL  \cite{wan2024dell}     & 63.79	& 62.26	&& 63.57&	62.19	&&{\ul 62.38}&{\ul 	61.57} \\
RAEmo  \cite{liu2024raemollm}    & {\ul 64.79}&	63.86&&	{\ul 64.35}&	{\ul 63.76}&&	61.52	&60.58 \\
\rowcolor{gray!50} MARO (ours) & \textbf{65.04}&	\textbf{64.27}&&	\textbf{65.28}&	\textbf{64.28}&&	\textbf{63.65}&	\textbf{62.88} \\
\midrule \midrule
\multicolumn{1}{c|}{\multirow{2}{*}{\textbf{Method}}}  & \multicolumn{2}{c}{\textbf{Ottawa Shooting}} && \multicolumn{2}{c}{\textbf{Sydney Siege}} && \multicolumn{2}{c}{\textbf{Avg.}} \\
\cmidrule(l){2-3} \cmidrule(l){5-6} \cmidrule(l){8-9}
& \textbf{Acc.} & \textbf{F1}  &&\textbf{Acc.} & \textbf{F1}  && \textbf{Acc.} & \textbf{F1} \\
\midrule
TELLER  \cite{liu-etal-2024-teller}  &61.06&	60.72&&	56.91	&56.77&&	61.30&	60.87\\
DELL  \cite{wan2024dell}    & 61.39&	61.28&&	56.82&	56.08&&	61.59	&60.68 \\
RAEmo  \cite{liu2024raemollm}    &{\ul 63.15}&
 {\ul 62.36}&&	{\ul 57.35}&	{\ul 57.08}	&& {\ul 62.23}&	{\ul 61.53}\\
\rowcolor{gray!50} 
MARO (ours) & \textbf{64.02}&	\textbf{63.66}&&	\textbf{59.25}&	\textbf{58.85}&&	\textbf{63.45}&	\textbf{62.79} \\
\bottomrule
\end{tabular}
\caption{{Performance comparison on PHEME using LLaMA-3.1-405B as the
underlying model.}}
\label{pheme_405B}
\end{table*}

\subsection{More Underlying Models}
\label{more underlying models}
We replace the underlying models for MARO and the strong baselines with LLaMA-3.1-405B, LLaMA-3.1-8B, and Claude-3.5-Sonnet. As shown in Tables \ref{10}-\ref{table wamt 8B}, MARO's performance remains superior to these baselines across different underlying models, demonstrating its effectiveness.
\begin{table*}[ht!]
\centering
\small
\setlength{\tabcolsep}{11.5pt}
\begin{tabular}{l|cccccccc}
\toprule
\multicolumn{1}{c|}{\multirow{2}{*}{\textbf{Method}}}  & \multicolumn{2}{c}{\textbf{Charlie Hebdo}} &&  \multicolumn{2}{c}{\textbf{Ferguson}}  && \multicolumn{2}{c}{\textbf{Germanwings Crash }} \\
\cmidrule(l){2-3} \cmidrule(l){5-6} \cmidrule(l){8-9}
& \textbf{Acc.} & \textbf{F1}  &&\textbf{Acc.} & \textbf{F1}  && \textbf{Acc.} & \textbf{F1} \\
\midrule
TELLER  \cite{liu-etal-2024-teller}   & 67.54	& 66.35&&	66.27	&{\ul 66.91}	&& {\ul 62.28}	&{\ul 61.57}\\
DELL  \cite{wan2024dell}    & 68.05&	\textbf{66.87}&&	{\ul 67.54}&	66.89	&&61.05&	60.59\\
RAEmo   \cite{liu2024raemollm}   & {\ul 68.26}&	63.26	&&67.37	& 65.37	&&61.26	&60.85\\
\rowcolor{gray!50} 
MARO (ours) & \textbf{73.31}	& {\ul 66.52}&&	\textbf{70.85}	&\textbf{67.01}	&&\textbf{63.06}&	\textbf{63.11}\\
\midrule
\midrule
\multicolumn{1}{c|}{\multirow{2}{*}{\textbf{Method}}}  & \multicolumn{2}{c}{\textbf{Ottawa Shooting}} &&  \multicolumn{2}{c}{\textbf{Sydney Siege}}  && \multicolumn{2}{c}{\textbf{Avg.}} \\ 
\cmidrule(l){2-3} \cmidrule(l){5-6} \cmidrule(l){8-9}
& \textbf{Acc.} & \textbf{F1}  &&\textbf{Acc.} & \textbf{F1}  && \textbf{Acc.} & \textbf{F1} \\
\midrule
TELLER \cite{liu-etal-2024-teller}   & 50.25 &	49.67&&	{\ul 49.26}	&\textbf{50.35}	&&59.12&	{\ul 58.97} \\
DELL  \cite{wan2024dell}     & {\ul 51.52}&	{\ul 50.67}&&	48.23	& 47.52&&	{\ul 59.28} &	58.51 \\
RAEmo  \cite{liu2024raemollm}     & 49.28	& 48.05&&	47.26	& 46.35&&	58.69	&56.78\\
\rowcolor{gray!50} 
MARO (ours) & \textbf{52.01}&	\textbf{51.88}&&	\textbf{50.38}&	{\ul 48.93}	&& \textbf{61.92} &	\textbf{59.49}\\
\bottomrule
\end{tabular}
\caption{{Performance comparison on PHEME using LLaMA-3.1-8B as the underlying model.}}
\label{pheme_8B}
\end{table*}
\begin{table*}[!t]
    \centering
    \small
\renewcommand{\arraystretch}{1.5}
    \begin{tabular}{p{14cm}c}\toprule
\multicolumn{1}{c}{Decision Rule}&\textbf{Acc.}\\ \midrule
Analyze the credibility of the news outlet and its fact-checking history regarding the social media event. If the news outlet has a history of spreading misinformation, output "1" as fake news; if the news outlet is known for credible reporting, output "0" as real news. Output requirements: - Output format: judgment: $<$'1' represents fake-news, '0' represents real-news$>$&55.31\\
&\\
Evaluate the cross-referencing of multiple reliable sources to verify the accuracy and credibility of the information presented in the news item. If the information is corroborated by multiple reputable sources, output "0" as real news; if there are conflicting reports or lack of consensus among sources, output "1" as fake news. Output requirements: - Output format: judgment: $<$'1' represents fake-news, '0' represents real-news$>$&62.52\\
&\\
Utilize sentiment analysis and social media monitoring to assess public reactions and discussions surrounding the social media event. If a large portion of the online community expresses skepticism or disbelief in the news item, output "1" as fake news; if the overall sentiment is positive and supportive of the news, output "0" as real news.    Output requirements:      - Output format: judgment: $<$'1' represents fake-news, '0' represents real-news$>$ &65.46\\
&\\
Evaluate the linguistic features and narrative structure of the news item to determine the level of bias and sensationalism in the reporting. If the article contains emotionally charged language, subjective opinions presented as facts, or sensationalized headlines, output "1" as fake news; if the article maintains a neutral tone, presents facts objectively, and avoids sensationalism, output "0" as real news. Output requirements: - Output format: judgment: $<$'1' represents fake-news, '0' represents real-news$>$ &65.68\\
&\\
Examine the consistency of the news item with verified data and expert opinions related to the social media event. If the news item aligns with established facts and expert analysis, output "0" as real news; if the news item contradicts verified data or expert opinions, output "1" as fake news. Output requirements: - Output format: judgment: $<$'1' represents fake-news, '0' represents real-news$>$ &68.39\\ \bottomrule
    \end{tabular}
    \caption{An example of the decision rule optimization process on Weibo21.}
\label{case study weibo21}
\end{table*}

\subsection{More Datasets}
\label{More Datasets}
We also conduct experiments on PHEME \cite{buntain2017automatically}, which is an {English} rumor detection dataset containing posts and comments related to five breaking events. Table \ref{pheme dataset} shows the statistics of PHEME. Similar to the above experiments, we conduct cross-event misinformation detection experiments on each event. As shown in Tables \ref{pheme_gpt3.5}-\ref{pheme_8B}, compared with the strong baselines, MARO still achieves the best performance on PHEME, demonstrating its effectiveness.

\section{Case Study}
\label{case stydy}
Table \ref{case study weibo21} shows an example of the decision rule optimization process. The left side of the table shows the generated decision rules, while the right side shows their validation accuracy. We can observe that decision rules with higher accuracy generally have stronger applicability.

\end{document}